\pdfoutput=1

\documentclass[11pt]{article}

\usepackage[]{acl}

\usepackage{times}
\usepackage{latexsym}

\usepackage{amsmath}
\usepackage{hyperref}
\usepackage{cleveref}

\usepackage[T1]{fontenc}

\usepackage[utf8]{inputenc}

\usepackage{microtype}

\usepackage{inconsolata}

\usepackage{multirow}
\usepackage{amsmath}
\usepackage{caption}
\usepackage{subcaption}
\usepackage{booktabs}
\usepackage{tabularx}
\usepackage{graphicx}
\usepackage{colortbl}
\usepackage{lipsum}
\usepackage{appendix}
\usepackage{amssymb}
\usepackage{amsmath}

\usepackage{algorithm}
\usepackage{algpseudocode}
\usepackage{subcaption}

\usepackage{color,soul}
\usepackage{xcolor}
\usepackage[colorinlistoftodos,prependcaption,textsize=tiny]{todonotes}
\definecolor{olivegreen}{RGB}{107,142,35}
\definecolor{lightolivegreen}{RGB}{157,192,105}

\newcommand{\xdetox}{\texttt{DetoxLLM}}

\definecolor{mygreen}{RGB}{0, 128, 0}

\title{\includegraphics[scale=0.03]{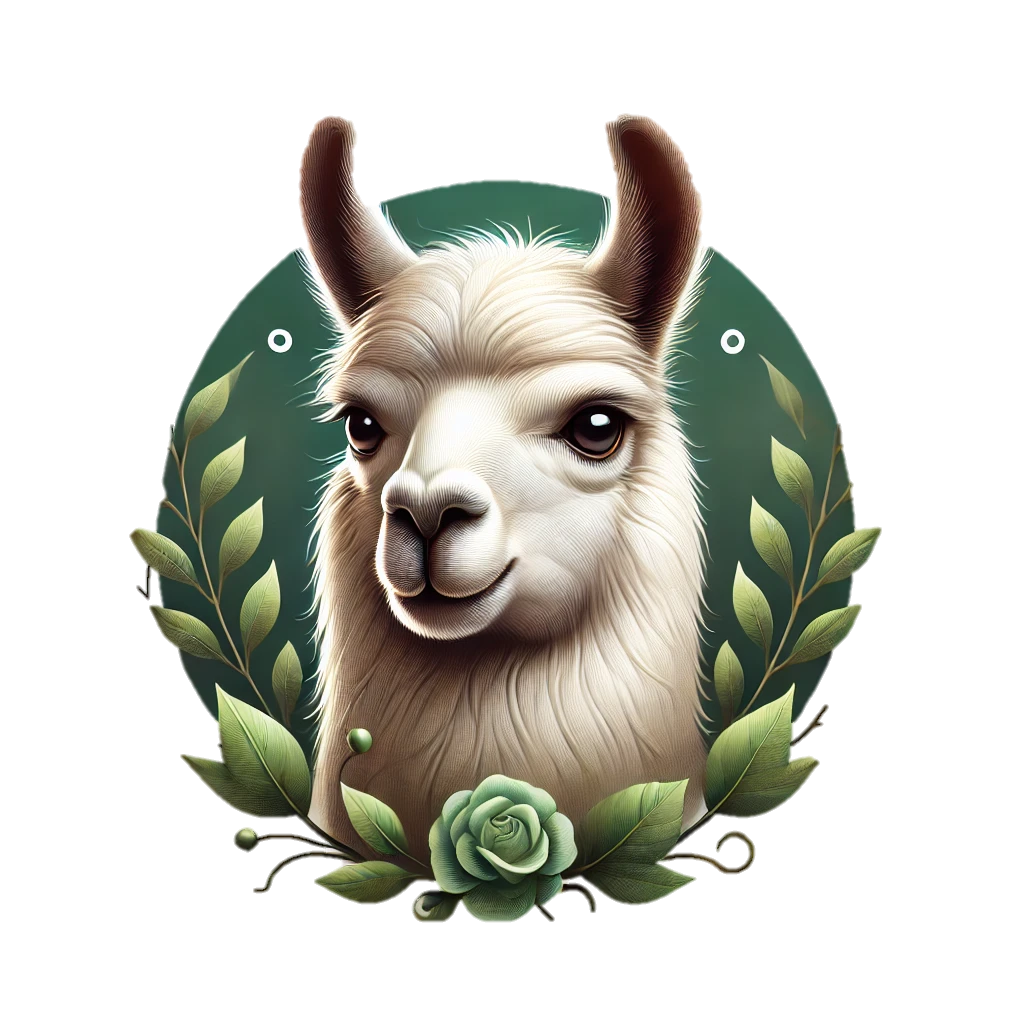} \xdetox:\\A Framework for Detoxification with Explanations\\{\small \textcolor{red}{\textit{Warning! This paper contains examples of toxic language}}}} 

\author{Md Tawkat Islam Khondaker$^{\spadesuit}$~~ {Muhammad Abdul-Mageed$^{\spadesuit}$$^\diamondsuit$}~~ {Laks V.S. Lakshmanan$^{\spadesuit}$}
\\\\ 
\normalsize $^{\spadesuit}$The University of British Columbia, $^\diamondsuit$MBZUAI \& Invertible AI  \\
\\
\texttt{\{tawkat@cs.,muhammad.mageed@,laks@cs.\}ubc.ca}
}

\newcommand{\LL}[1]{#1}

\newcommand{\eat}[1]{}

\newcommand{\modification}[1]{#1}

\begin{document}
\maketitle
\begin{abstract}
Prior works on  detoxification  are scattered in the  sense that they do not cover all aspects of detoxification needed in a real-world scenario. Notably, prior works restrict the task of developing detoxification models to only a seen subset of platforms, leaving the question of how the models would perform on  unseen platforms unexplored. Additionally, these works do not address non-detoxifiability, a phenomenon whereby  the toxic text cannot be detoxified without altering the meaning. We propose~\xdetox\footnote{\href{https://huggingface.co/UBC-NLP/DetoxLLM-7B}{UBC-NLP/DetoxLLM-7B}}, the first comprehensive \LL{end-to-end} detoxification framework, which attempts to alleviate the aforementioned limitations. We first introduce a cross-platform pseudo-parallel corpus applying multi-step data processing and generation strategies leveraging \texttt{ChatGPT}. We then train a suite of detoxification models with our cross-platform corpus. We show that our detoxification models outperform the SoTA model trained with human-annotated parallel corpus. We further introduce explanation to \LL{promote}  transparency and trustworthiness.~\xdetox~additionally \LL{offers} a unique paraphrase detector especially dedicated for the detoxification task to tackle \LL{the non-detoxifiable cases.} Through experimental analysis, we demonstrate the effectiveness of our cross-platform corpus and the robustness of \xdetox~against adversarial toxicity.


\end{abstract}

\section{Introduction}
\begin{figure}[t]
    \centering
  \includegraphics[width=0.8\columnwidth]{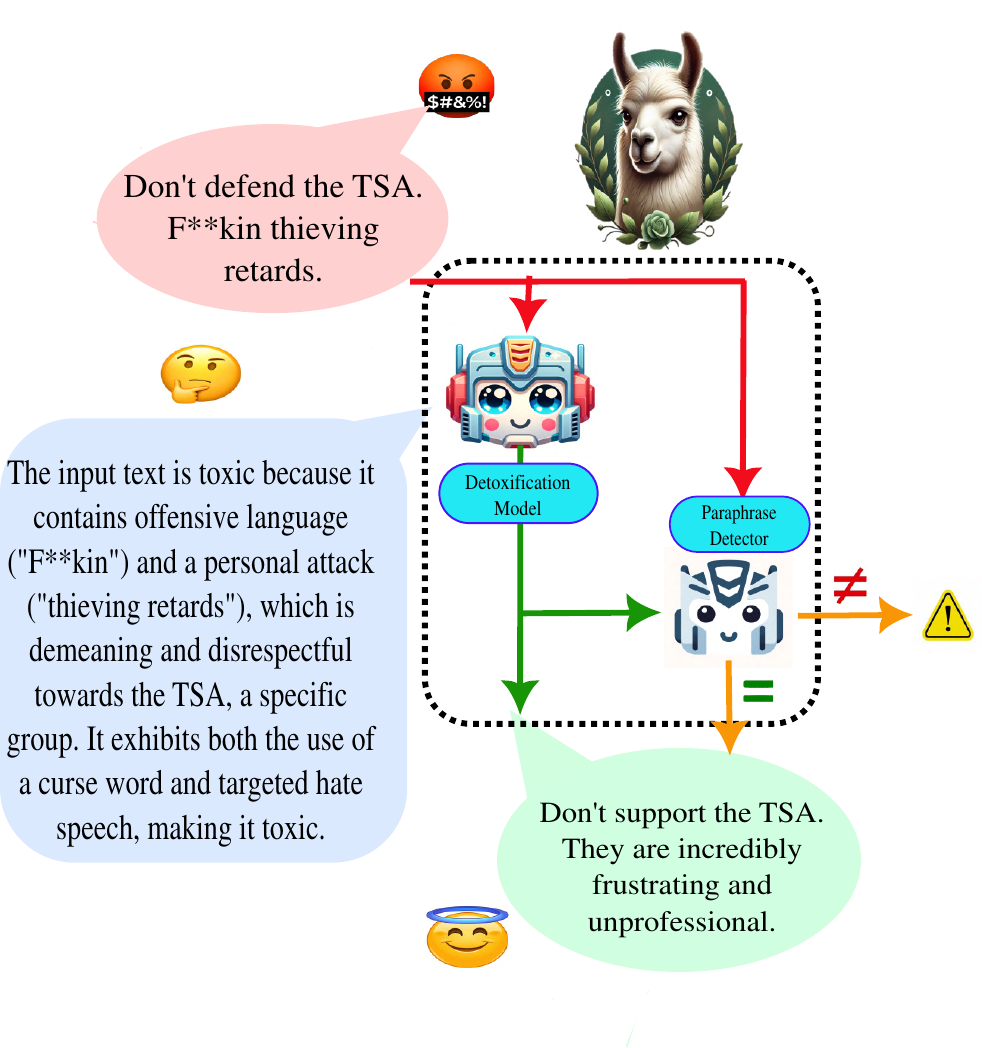}
  \caption{\label{fig:xdetox_intro}
  Workflow of \xdetox~framework. The framework will take a \colorbox{red!30}{toxic} input. The detoxification model will generate the \colorbox{blue!15}{explanation} of why the input is toxic, as well as a \colorbox{green!30}{non-toxic} version. The paraphrase detector will analyze the semantic similarity of the toxic and non-toxic pair and generate a \colorbox{yellow!30}{warning} if the pair is not semantically equivalent (an illustration of non-detoxifiable case is depicted in Appendix~\ref{appendix:illustration_for_nondetoxifiability}).
  }
\end{figure}

The term \textit{toxic language} is usually used to refer to any form of offensive or hateful speech~\citep{laugier-etal-2021-civil,fortuna-toxic-lrec}; specifically,  toxic or abusive language is defined as any form of microaggression, condescension, harassment, hate speech, trolling, and the like~\citep{jurgens}. Use of toxic language online has been a significant issue over the years. Although a plethora of works have explored the task of toxicity detection, the task remains challenging  due to its evolving nature~\cite{davidson,muller_2017,williams}. In addition, the linguistic variation in how toxicity manifests itself across different platforms~\cite{karan,swamy,salminen_2020} poses  a standing challenge for toxicity detection. \LL{Furthermore,} the \LL{task of detecting} toxic language, \LL{taken literally,}  can only offer deletion of \LL{toxic} text. A more comprehensive approach to dealing with toxic text would be to rewrite the text to keep the useful content intact and eliminate toxicity, \LL{a task} known as \textit{detoxification}~\citep{logacheva-etal-2022-paradetox}. Several works~\citep{dos-santos-2018,dale-etal-2021-skoltechnlp} have already explored the idea of detoxification. More recently,~\citet{logacheva-etal-2022-paradetox} propose \textit{ParaDetox}, the first detoxification model developed with a crowd-sourced parallel corpus, \LL{which  outperforms}  the unsupervised competitors in the detoxification task. 

\LL{Unfortunately},  prior works focus on only a particular subproblem when tackling detoxification, overlooking  other important aspects of the problem, detailed below. \textbf{(1)} previous works~\citep{dos-santos-2018,dale-etal-2021-skoltechnlp} have only explored the idea of in-platform detoxification, i.e., the models are trained and tested on the same platforms, as opposed to  cross-platform detoxification, where the training platforms (e.g., Wikipedia, Reddit) are disjoint from the testing platforms (e.g., Facebook, Youtube). As a result, how the detoxification models would \LL{perform on} different platforms and \LL{cope with the linguistic variation present across platforms} is still an unexplored territory. \textbf{(2)} Secondly, prior works do not justify why a given input is \LL{found to be} toxic~\citep{logacheva-etal-2022-paradetox}. When we intend to deploy a detoxification model in the real-world, we also need to explain \textit{why} we are altering a given text. Therefore, we intend to incorporate explanation as a part of our system design to assist  users engage in healthy communication, thus enhancing transparency and the credibility of the system itself. \textbf{(3)}  Current works do not \LL{properly} tackle \textit{non-detoxifiability}, a phenomenon whereby a toxic text cannot be detoxified without altering the meaning. As a consequence, deploying a system without handling non-detoxifiability can make it ineffective in real-life scenarios.  \textbf{(4)} Finally, even with the advent of generalized large language models (LLMs)~\citep{alpaca,vicuna,jiang2023mistral,qwen2.5,team2024gemma,abdin2024phi}, the detoxification task remains challenging since instruction-tuned LLMs often refuse to respond to toxic input due to their safety \LL{requirements}~\citep{llama2} (see \S\ref{sec:performance_instruction_llm}). 


In this work, we offer a comprehensive and realistic detoxification framework that resolves issues with prior works on detoxification. More specifically, we introduce \xdetox, the first end-to-end framework for the detoxification task (Figure~\ref{fig:xdetox_intro}), focusing on piecing together \LL{our}  solutions for all issues discussed above. Given a toxic text, our detoxification model will first analyze and provide an explanation as to why the input is \LL{found} toxic. Then, the model will \LL{attempt} to detoxify and \LL{output} the non-toxic version of the input. \LL{Unlike} prior works~\citep{dale-etal-2021-skoltechnlp,logacheva-etal-2022-paradetox}, we additionally incorporate a dedicated paraphrase detector in our framework to tackle the cases of \textit{non-detoxifiability}. If the input is non-detoxifiable,~\xdetox~will prompt an additional \LL{warning} to the user regarding \LL{possible meaning alteration in the text.} To train our detoxification models on cross-platform corpus,  we first collect a wide array of annotated toxic and non-toxic data from different existing works. We then employ \texttt{ChatGPT}\footnote{gpt-3.5-turbo from June, 2023.}~\citep{openai-chatgpt} through a meticulous prompt engineering approach to build a pseudo-parallel corpus.

Our contributions can be summarized as follows:
\begin{enumerate}
    \item We propose \xdetox, the first detoxification framework that tackles toxic language  across different platforms as well as handles non-detoxifiability while providing explanation for the toxic input.
    
    \item We develop the first cross-platform pseudo-parallel detoxification corpus with multi-step data processing and prompt engineering.
    
    \item We empirically evaluate and compare our detoxification models against SoTA baselines. Our experiments show that~\xdetox~outperforms SoTA in cross-platform detoxification, and our detoxification model \textit{CoT-expl} LLaMA of~\xdetox~achieves the best performance.


    \item We train a unique paraphrase detector \LL{tailored} for the detoxification task in order to handle the cases of non-detoxifiability. Our comparative evaluation against the SoTA paraphrase detectors clearly illustrates the necessity of such a specialized detector dedicated to the detoxification task.
    
    \item We conduct an extensive experimental analysis to demonstrate the effectiveness of our cross-platform data as well as the robustness of \xdetox~against implicit and token-level adversarial toxicity.


\end{enumerate}


\begin{figure*}[t]
  \centering
  \includegraphics[width=0.9\textwidth]{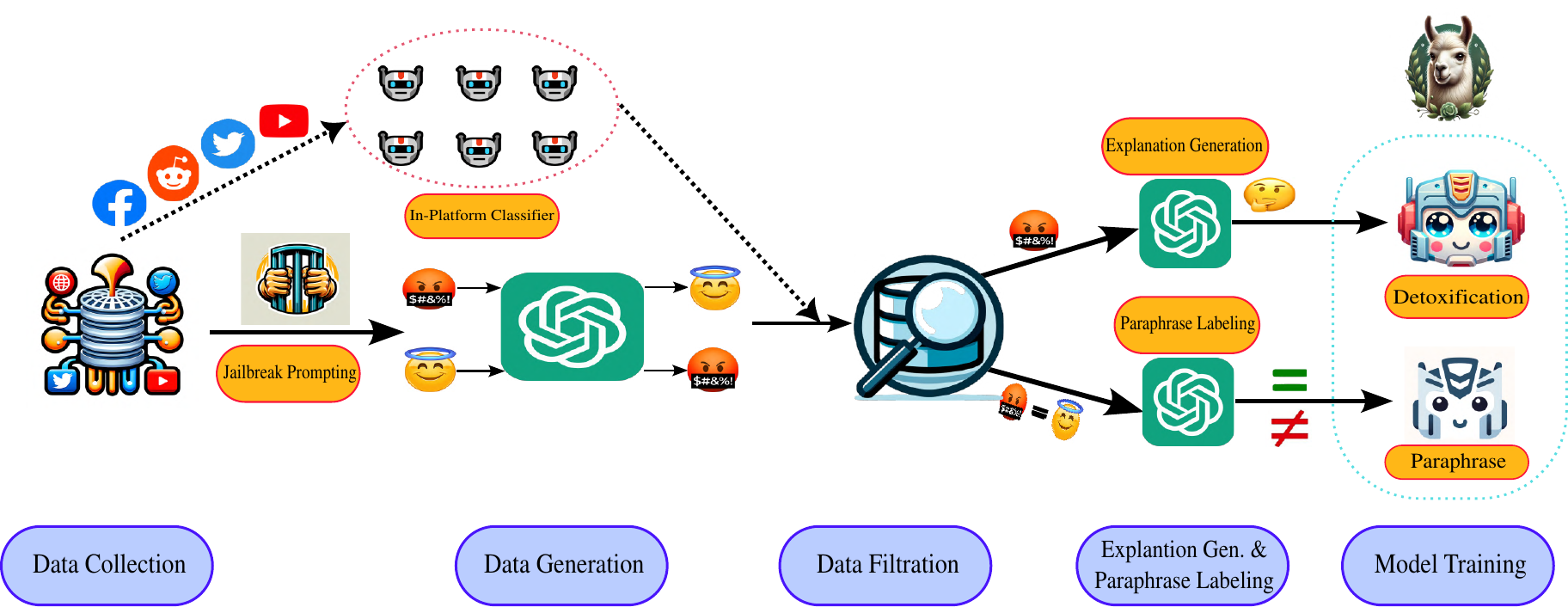}
  \caption{\label{fig:methodology}
  Overall methodology of \xdetox. Initially, we collect the toxicity corpus from multiple platforms (\S\ref{sec:data_collection}). Then, we generate texts of opposite classes (\S\ref{sec:data_generation}). We filter out ambiguous data (\S\ref{sec:data_filtration}). After that, we generate explanation and paraphrase labels (\S\ref{sec:explanation_paraphrase}). Finally, we train the detoxification and the paraphrase detection models (\S\ref{sec:model_training}).
  }
\end{figure*}

\section{Related Works}
\label{sec:Related_Works}
Over the years, several works have  studied  abusive language detection~\citep{founta,davidson,golbeck,waseem}. The task of text style transfer (TST) has also been explored in the field of NLP due to its wide range of applications~\citep{shen-style-2017,rao-formality-2018,patel-authorship-2022,mukherjee-politeness-2023}. Notably, studies like \citet{reif-llm-2022,pu-llm-2023} show the effectiveness of LLMs for parallel data generation and style transfer tasks. Inspired by these works, we resort to \LL{use} LLMs in our work for pseudo-parallel dataset creation and consequently distill the knowledge in comparatively smaller language models. We provide a detailed account of related works on abusive language detection and TST in~Appendix~\ref{appendix:related_works}.

\noindent \textbf{Detoxification} is formulated as style transfer from toxic to neutral and non-toxic style~\citep{logacheva-etal-2022-paradetox,pour-etal-2023-count}. Prior works like~\citet{dos-santos-2018} and~\citet{laugier-etal-2021-civil} create their own detoxification corpus from Reddit and Jigsaw~\citep{jigsaw}, respectively. \citet{dale-etal-2021-skoltechnlp} employ style-trained language models to guide a paraphraser preserve the content and remove toxicity. The authors further use the masked language modeling strategy of BERT~\citep{bert} to replace the toxic tokens with its non-toxic alternatives.~\citet{logacheva-etal-2022-paradetox} develop a human-annotated parallel corpus from Jigsaw, X (formerly known as Twitter), and Reddit. The authors train a BART~\citep{bart} model on this parallel corpus and achieve the SoTA performance on detoxification, showing the importance of  high quality parallel data. Recently, \citet{dementieva-2023-detoxfication} propose cross-lingual detoxification through simultaneous text translation and detoxification.

However, none of the prior works explore the idea of cross-platform detoxification potentially due to the scarcity of parallel data. This research gap motivates  our work on this particular subproblem.

\section{Proposed Methodology}
\label{sec:Proposed_Methodology}

We present our methodology in Figure~\ref{fig:methodology} (please see the caption for the overview). Now, we describe each component of our cross-platform detoxification \LL{framework}.

\subsection{Data Collection}
\label{sec:data_collection}

To create a cross-platform parallel detoxification corpus, we first compile datasets from a wide range of platforms. We collect the sources of the datasets primarily from~\citet{unified_datasets} and~\citet{hatespeech_datasets}. Table~\ref{table:datasets} provides details of these datasets.

\begin{table}[h]
\centering
\footnotesize  
\scalebox{0.6}{
\begin{tabular}{llcll}
\hline
\textbf{Dataset} & \textbf{Platform}                                              & \textbf{Source}                                                                                                                                                                                                                                                                                                                                                                                                                 & \textbf{Toxic/Normal}  &  \textbf{Original/Filtered}\\ \hline
wiki             & Wikipedia                                                      & \citet{wulczyn}                                                                                                                                                                                                                                                                                                                                                                                            & 14,880 / 117,935   & 3,000 / 2,153                  \\

twitter          & Twitter                                                        & \textit{Multiple*} & 77,656 / 55,159  & 3,000 / 2,337
\\

fb-yt           & \begin{tabular}[c]{@{}c@{}}Facebook \\\& Youtube\end{tabular}  & \citet{salminen}                                                                                                                                                                                                                                                                                                                                                                                           & 2,364 / 858        &   2,897 / 1,901                  \\
stormfront       & Stormfront                                                     & \citet{gibert}                                                                                                                                                                                                                                                                                                                                                                                            & 1,364 / 9,507      &   3,000 / 2,511                   \\
fox              & Fox News                                                      & \citet{gao}                                                                                                                                                                                                                                                                                                                                                                                                 & 435 / 1,093      &    1,104 /  831               \\

reddit           & Reddit                                                         & \citet{qian}                                                                                                                                                                                                                                                                                                                                                                                               & 2,511 / 11,073       &   3,000 / 2,222                 \\
convAI           & \begin{tabular}[c]{@{}c@{}}ELIZA \&\\ CarbonBot\end{tabular}   & \citet{curry}                                                                                                                                                                                                                                                                                                                                                                                             & 128 / 725           &   650 / 552                \\
hateCheck        & \begin{tabular}[c]{@{}c@{}}Synthetic.\\ Generated\end{tabular} & \citet{rottger}                                                                                                                                                                                                                                                                                                                                                                                          & 2,563 / 1,165      &   2,741 / 1,398                   \\
gab              & Gab                                                            & \citet{qian}                                                                                                                                                                                                                                                                                                                                                                                               & 15,270 / 656       &   3,000 / 2,151              \\
yt\_reddit       & \begin{tabular}[c]{@{}c@{}}Youtube \\\& Reddit\end{tabular}    & \citet{mollas}                                                                                                                                                                                                                                                                                                                                                                                           & 163 / 163      &   222 / 156                     \\ \hline
\end{tabular}
}
\caption{\label{table:datasets}
List of experimental datasets with \LL{varying toxic/normal ratio and} the corresponding platforms. \modification{We further show the original/filtered ratio after applying data filtration process} (\S\ref{sec:data_filtration}). \textbf{*} \emph{Twitter} dataset is collected from~\citet{waseem},~\citet{davidson},~\citet{jha},~\citet{elsherief},~\citet{founta},~\citet{mathur},~\citet{basile},~\citet{mandl},~\citet{ousidhoum}, and~\citet{olid}.
}
\end{table}

Some datasets in Table~\ref{table:datasets} provide multi-class toxicity labeling such as \textit{hate}, \textit{offensive}, \textit{accusation}. We label all of these classes as \textit{toxic} and transform all the dataset into binary classification (\textit{toxic} vs.  \textit{non-toxic}). To keep the cost manageable and avoid overfitting, we randomly select at most $3,000$ samples from each dataset.

\subsection{Data Generation through Jailbreaking}
\label{sec:data_generation}

\begin{figure*}[t]
     \centering
     \begin{subfigure}[b]{0.32\textwidth}
         \centering
         \includegraphics[width=\textwidth]{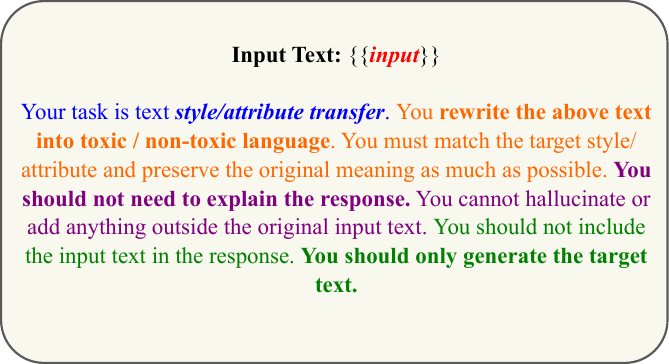}
         \caption{Parallel Data Generation}
         \label{fig:parallel_data_generation}
     \end{subfigure}
    \hfill
     \begin{subfigure}[b]{0.32\textwidth}
         \centering
         \includegraphics[width=\textwidth]{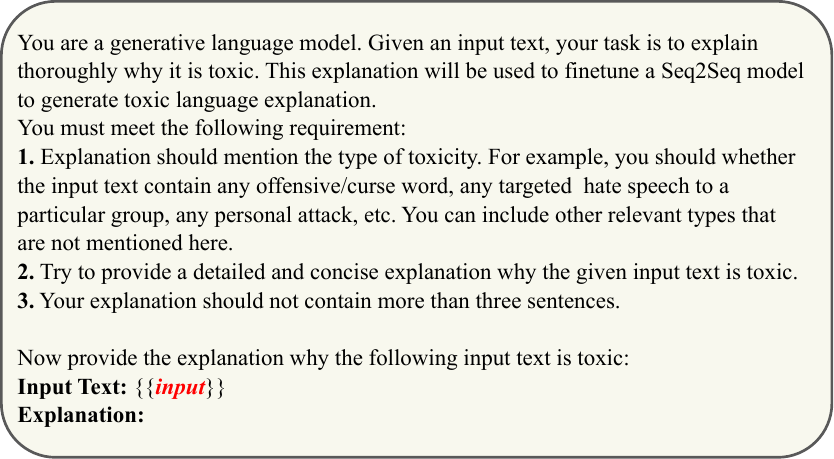}
         \caption{Explanation Generation}
         \label{fig:explanation_generation}
     \end{subfigure}
    \hfill
     \begin{subfigure}[b]{0.32\textwidth}
         \centering
         \includegraphics[width=\textwidth]{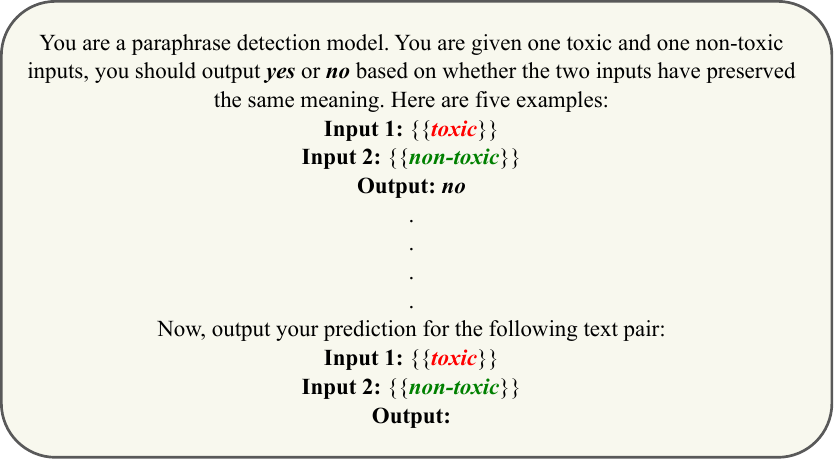}
         \caption{Paraphrase Labeling}
         \label{fig:paraphrase_labeling}
     \end{subfigure}
        \caption{Prompt design for toxic, non-toxic parallel data generation (\S\ref{sec:data_generation}), explanation generation,and  paraphrase labeling (\S\ref{sec:explanation_paraphrase}) with \texttt{ChatGPT}.}
        \label{fig:prompt_design}
\end{figure*}

To train our models on cross-platform detoxification, we require parallel non-toxic as well as toxic data. While \texttt{ChatGPT}~\citep{openai-chatgpt} is developed with safety mechanisms to restrict the model's behavior to be safe~\citep{openai-moderation}, this restriction can be manipulated through careful engineering of prompts,  \LL{a process} known as \textit{jailbreaking}~\citep{Li2023MultistepJP,jailbreak_chat}. In the context of language modeling, jailbreaking refers to the process of circumventing the restrictions placed on models~\citep{Liu2023JailbreakingCV}. Hence, we apply jailbreaking to design a prompt that can  exploit \texttt{ChatGPT} to generate parallel toxic text given  non-toxic version and vice versa. Our jailbreaking prompt includes the following components: \textbf{(1)} We first deliver toxic/non-toxic \textcolor{red}{\textit{input}} to the model, \textbf{(\{\{ input \}\})}. \textbf{(2)} We then set the \textcolor{blue}{\textit{task}} of the model (e.g., \textit{style/attribute transfer}). \textbf{(3)} We provide the \textcolor{orange}{\textit{objective}}  of the model (e.g., \textit{provide the parallel text of opposite label for the input text}). \textbf{(4)} We add explicit \textcolor{violet}{\textit{constraints}} to the model's generation (e.g., \textit{Do not explain or hallucinate}). \textbf{(5)} Finally, we define what the expected \textcolor{mygreen}{\textit{response format}} of the model \LL{is} (e.g., \textit{Do not include input text in response}). We present the template of our designed prompt in Figure~\ref{fig:parallel_data_generation}.

\subsection{Data Filtration}
\label{sec:data_filtration}

Distinguishing between types of toxic text e.g., offensive language and hate speech, is often deemed subjective~\citep{sap,wilds}: a text labeled \textit{non-toxic} on one platform may be considered \textit{toxic} on another. To avoid cross-platform ambiguity, we first train in-house platform-specific toxicity classifiers on six datasets (\emph{fb-yt}, \emph{fox}, \emph{twitter}, \emph{stormfront}, \emph{wiki}, \emph{hateCheck}) separately. Then we predict the toxicity of the parallel data in our corpus. We only select those samples where at least one classifier predicts the source text (a.k.a. toxic text) as \textit{toxic} \textbf{AND} all the classifiers predict the target text (a.k.a. non-toxic text) as \textit{non-toxic}. In other words, we filter out any toxic sample that is predicted to be \textit{non-toxic} by all the classifiers and we also filter out any non-toxic sample that is predicted to be \textit{toxic} by at least one classifier. Finally, \underline{to experiment with cross-platform detoxification,} we only select \emph{wiki}, \emph{reddit},  and \emph{twitter} for training to keep the training platforms compatiable with~\citet{logacheva-etal-2022-paradetox} for fair comaprison. We show the number of samples for each platform before (original) and after (filtered) the data filtration process in Table~\ref{table:datasets}.

\subsection{Explanation and Paraphrase Acquisition}
\label{sec:explanation_paraphrase}

To generate explanation using the models and train our models with Chain-of-Thought (CoT) prompting~\citep{wei2022chain}, we further generate the explanation of toxicity from \texttt{ChatGPT}. Hence, we prompt \texttt{ChatGPT} with the toxic texts from the filtered dataset (Section~\ref{sec:data_filtration}) and ask it why the given text is deemed  toxic (Figure~\ref{fig:explanation_generation}). Similar to Section~\ref{sec:data_generation}, we design a specific prompt instructing \texttt{ChatGPT} to describe the type of toxicity (e.g., personal or group attack). We also constrain \texttt{ChatGPT} to explain in at most three sentences. For paraphrase labeling, we first pass five manually labeled few-shot examples. Then, we provide both toxic and non-toxic parallel data to \texttt{ChatGPT} and prompt the model to analyze whether the text-pair is semantically similar (Figure~\ref{fig:paraphrase_labeling}). We provide samples of our cross-platform dataset in Appendix~\ref{appendix:cross_platform_dataset_sample}.

\subsection{Model Training}
\label{sec:model_training}

We finetune both encoder-decoder and decoder-only models for detoxification. For encoder-decoder, we use BART~\citep{bart} and T5~\citep{t5} with their base architectures. For decoder-only models, we finetune LLaMA-2~\citep{llama2} with $7$B parameters. We train the models with direct toxic to non-toxic finetuning (\underline{vanilla}) as well as prompt $\oplus$ toxic to non-toxic fientuning (\underline{prompt}), where we first give a prompt instructing the model to perform detoxification style transfer. We further employ CoT finetuning where the models are first instructed to generate explanation and then based on the toxic input incorporated with explanation, models generate the non-toxic version (\underline{CoT-expl}). (See Figure~\ref{fig:CoT_prompt_design} for prompt template). To detect paraphrasing of a toxic and a non-toxic text pair, we further finetune a BERT~\citep{bert} classifier with the paraphrase labels generated by \texttt{ChatGPT}.

\modification{We note that  unlike  prior work~\citep{logacheva-etal-2022-paradetox} which excludes non-detoxifiable texts, we generate non-toxic (not meaning-preserving) outputs from these toxic texts. Therefore, upon training our detoxification models with such data, the models will learn to produce non-toxic (but not meaning-preserving) texts. Then the source-target pair will be passed to the paraphrase detector. Consequently, the detector should label the pair as ``non-paraphrase", indicating the non-detoxifiability and prompting an additional warning (Figure~\ref{fig:xdetox_intro}).}

\section{Experiments}
\label{sec:experiments}
\textbf{Models Compared.} \textbf{(1)} \colorbox{red!15}{\textit{SoTA Baseline.}} ParaDetox, a BART-based model developed by~\citet{logacheva-etal-2022-paradetox} and LLaMA-2~\citep{llama2} model finetuned on ParaDetox. \textbf{(2)} \colorbox{orange!15}{\textit{*-DSS.}} BART and T5 models trained with SoTA distillation method proposed by~\citet{dss-Hsieh}. \textbf{(3)} \colorbox{yellow!15}{\textit{Instruction-tuned.}} Alpaca~\citep{alpaca}, LLaMA-2 (Chat), and Vicuna~\citep{vicuna}. We use the corresponding 7B versions. \textbf{(4)} \colorbox{mygreen!15}{\textit{Cross-Platform Models.}} Our suite of models (BART, T5, and LLaMA-2-7B) trained on the cross-platform datasets (\S\ref{sec:model_training}).

\noindent \textbf{Performance Metrics.} \textbf{(1)} \textit{Accuracy.} We compute accuracy of the models based on the percentage of non-toxic outputs identified by the same RoBERTa style classifier as~\citet{logacheva-etal-2022-paradetox}. We provide accuracy measured by our in-house platforms (\S\ref{sec:data_filtration}) in Appendix~\ref{appendix:accuracy_cross_platforms}. \textbf{(2)} \textit{BERTScore.} We use BERTScore with SimCSE~\citep{simcse} RoBERTa-large model to compute how the models preserve the semantic meaning. 
\modification{\textbf{(3)} \textit{Content Similarity.} Cosine similarity between the embeddings of the original text and the output computed with the model of~\citet{wieting-etal-2019-beyond}.} \textbf{(4)} \textit{Fluency.} Following~\citet{logacheva-etal-2022-paradetox}, we measure the percentage of fluent sentences identified by a RoBERTa-based classifier trained on the linguistic acceptability (CoLA) dataset~\citep{cola-warstadt}. \modification{\textbf{(5)} \textit{Joint Metric.} Multiplication of \textit{Accuracy}, \textit{Content Similarity}, and \textit{Fluency}, as proposed by~\citet{logacheva-etal-2022-paradetox}} \textbf{(6)} \textit{BLEU.} We compute the BLEU score between the input and the corresponding output.

We provide  detailed information on the experiments  including implementation details, baselines, and performance metrics in Appendix~\ref{appendix:experimental_details}.

\section{Results}
\label{sec:Results}

\begin{table*}[t]
\centering
\resizebox{\linewidth}{!}{
\begin{tabular}{l|cccccc|cccccc|cccccc|cccccc|}
\hline
                   & \multicolumn{6}{c|}{\cellcolor{cyan!20}\textbf{yt\_reddit}}                                                    & \multicolumn{6}{c|}{\cellcolor{red!20}\textbf{fb\_yt}}                                                                          & \multicolumn{6}{c|}{\cellcolor{orange!20}\textbf{fox news}}                                                      & \multicolumn{6}{c|}{\cellcolor{green!20}\textbf{Overall}}                                                       \\ \hline
\textbf{Model}     & \textbf{ACC} & \textbf{BS} & \textbf{SIM} & \textbf{FL} & \textbf{J} & \textbf{BL} & \textbf{ACC} & \textbf{BS} & \textbf{SIM} & \textbf{FL} & \textbf{J}                   & \textbf{BL} & \textbf{ACC} & \textbf{BS} & \textbf{SIM} & \textbf{FL} & \textbf{J} & \textbf{BL} & \textbf{ACC} & \textbf{BS} & \textbf{SIM} & \textbf{FL} & \textbf{J} & \textbf{BL} \\ \hline
\cellcolor{red!15}\textbf{ParaDetox}          & 44.00        & 97.43       & 88.47        & 76.00       & 29.58      & 27.52       & 79.00        & 95.50       & 79.04        & 93.00       & 58.07                        & 21.16       & 78.00        & 97.37       & 85.68        & 96.00       & 64.16      & 35.08       & 67.86        & 96.51       & 82.17        & 91.14       & 50.29      & 29.13       \\
\cellcolor{orange!15}\textbf{T5-DSS}            & 67.39        & 95.70       & 76.35        & 97.83       & 50.34      & 35.73       & 72.41        & 95.74       & 78.73        & 98.85       & 56.35                        & 41.97       & 73.63        & 95.07       & 76.20        & 94.51       & 53.03      & 38.26       & 68.33        & 95.55       & 77.01        & 96.72       & 50.89      & 38.88       \\
\cellcolor{orange!15}\textbf{BART-DSS}          & 82.61        & 93.61       & 62.19        & 93.48       & 48.03      & 39.32       & 94.25        & 93.78       & 68.85        & 98.85       & 64.14                        & 47.45       & 86.81        & 94.04       & 69.11        & 95.60       & 57.35      & 43.64       & 85.77        & 93.85       & 68.07        & 97.80       & 57.19      & 43.53       \\
\cellcolor{mygreen!15}\textbf{T5-V}        & 62.00        & 94.48       & 72.10        & 98.00       & 43.81      & 34.16       & 76.00        & \textbf{96.23}       & \textbf{87.24}        & 98.00       & \textbf{64.98}                        & 42.10       & 88.00        & 92.85       & 63.95        & 99.00       & 55.71      & 34.78       & 74.86        & 94.08       & 68.41        & \textbf{98.71}       & 50.66      & 36.91       \\
\cellcolor{mygreen!15}\textbf{T5-P}           & 70.00        & 91.37       & 55.49        & 94.00       & 36.51      & 32.93       & 80.00        & 93.97       & 77.38        & 99.00       & 61.28                        & 40.84       & 87.00        & 91.46       & 52.29        & 98.00       & 44.58      & 37.24       & 75.43        & 91.61       & 54.32        & 97.71       & 40.90      & 36.59       \\
\cellcolor{mygreen!15}\textbf{T5-CE}       & 67.39        & 89.21       & 37.81        & 97.83       & 24.93      & 32.35       & 78.16        & 89.69       & 40.79        & 95.40       & 30.41                        & 37.93       & 72.53        & 89.48       & 38.87        & 96.70       & 27.26      & 34.25       & 74.10        & 89.57       & 40.56        & 96.23       & 28.94      & 34.91       \\
\cellcolor{mygreen!15}\textbf{BART-V}      & 88.00        & 92.88       & 62.53        & 98.00       & \textbf{53.93}      & 38.14       & 96.00        & 94.48       & 80.88        & 99.00       & 76.87                        & 45.85       & 93.00        & 94.48       & 70.66        & \textbf{100.00}      & 65.71      & 41.50       & 88.71        & 93.60       & 65.94        & 98.14       & 57.92      & 40.06       \\
\cellcolor{mygreen!15}\textbf{BART-P}       & 74.00        & 91.04       & 52.70        & 98.00       & 38.22      & 36.77       & 89.00        & 92.97       & 74.27        & \textbf{100.00}      & 66.10                        & 44.11       & 92.00        & 91.67       & 53.60        & 99.00       & 48.82      & 39.77       & 83.00        & 91.32       & 52.24        & 97.86       & 43.22      & 38.99       \\
\cellcolor{mygreen!15}\textbf{BART-CE}     & 80.43        & 89.27       & 37.56        & \textbf{100.00}      & 30.21      & 37.39       & 89.66        & 89.34       & 38.68        & \textbf{100.00}      & 34.68                        & 38.58       & 89.01        & 88.91       & 35.51        & 96.70       & 30.56      & 35.76       & 87.29        & 89.23       & 38.05        & 98.59       & 32.73      & 36.78       \\
\cellcolor{yellow!15}\textbf{Alpaca}    & 43.48        & 84.86       & 18.79        & \textcolor{gray}{100.00}      & 8.17       & 9.27        & 51.72        & 84.13       & 22.87        & 97.70       & 11.56 & 8.52        & 59.34        & 84.57       & 16.29        & 94.51       & 9.14       & 7.19        & 49.33        & 84.76       & 17.57        & 96.70       & 8.39       & 8.35        \\
\cellcolor{yellow!15}\textbf{LLaMA-C} & \textcolor{gray}{100.00}       & 84.53       & 24.08        & 97.83       & 23.56      & 11.93       & 95.40        & 84.20       & 27.83        & \textcolor{gray}{100.00}      & 26.55 & 18.27       & \textcolor{gray}{97.80}        & 84.26       & 20.27        & \textcolor{gray}{100.00}      & 19.82      & 10.05       & \textcolor{gray}{97.94}        & 84.41       & 20.48        & 99.07       & 19.86      & 11.41       \\
\cellcolor{yellow!15}\textbf{Vicuna}    & 86.96        & 84.46       & 20.26        & \textcolor{gray}{100.00}      & 17.62      & 12.04       & 80.46        & 84.26       & 24.94        & 98.85       & 19.84 & 14.82       & 80.22        & 84.46       & 16.32        & 96.70       & 12.66      & 8.49        & 82.54        & 84.63       & 18.39        & 98.42       & 14.92      & 10.63       \\
\cellcolor{red!15}\textbf{LLaMA-PD}     & 56.39	& \textbf{98.22}	& \textbf{90.32}	& 97.57	& 49.69 & 31.33 & 82.23 & 97.67 & 89.45 & 97.57 & 71.77 & 26.88 & 83.71	& \textbf{97.55} & \textbf{88.54} & 97.98 & \textbf{72.62} & 43.51 & 73.16 & \textbf{96.89} & \textbf{84.52} & 98.17 &	\textbf{60.31} & 34.80   \\

\cellcolor{mygreen!15}\textbf{LLaMA-P}     & 84.78        & 91.13       & 50.86        & 97.83       & 42.18      & 49.39       & 96.55        & 91.99       & 57.24        & 97.70       & 53.99                        & \textbf{67.89}       & 93.41        & 92.04       & 53.64        & 97.80       & 49.00      & \textbf{60.71}       & 92.02        & 91.83       & 55.66        & 98.42       & 50.51      & 59.19       \\
\cellcolor{mygreen!15}\textbf{LLaMa-CE}    & \textbf{97.83}        & 83.61       & 55.70        & 97.83       & 53.31      & \textbf{52.98}       & \textbf{98.85}        & 86.65       & 61.52        & 97.70       & 59.41                        & 67.54       & \textbf{95.60}        & 87.23       & 57.84        & 98.90       & 54.69      & 58.44       & \textbf{95.94}        & 88.22       & 58.05        & 98.42       & \textbf{54.82}      & \textbf{59.33}       \\ \hline
\end{tabular}
}
\caption{
\label{table:result_cross_platform}
Performance of the models on cross-platform datasets. \modification{We provide the performances on the rest of the platforms in Appendix~\ref{appendix:performance_other_cross_platforms}.} \textbf{Acc} = percentage of non- toxic outputs identified by a style classifier, \textbf{BS} = BERTScore, \textbf{Sim} = Content Similarity, \textbf{Fl} = Fluency, \textbf{J} = Joint Metric, \textbf{BL} = BLEU Score. \textbf{V} = Vanilla, \textbf{P} = Prompt, \textbf{PD} = ParaDetox-finetuned, \textbf{CE} = CoT-expl, \textbf{C} = Chat. \textbf{Bold} font represents the best performance for a particular metric. We separately show the best performance of the \colorbox{yellow!15}{instruction-tuned models} in \textcolor{gray}{gray} due to their inability to detoxification (Section~\ref{sec:performance_instruction_llm}).
}
\end{table*}

\noindent \textbf{Overview.} We present the performance of the models on cross-platform detoxification in Table~\ref{table:result_cross_platform}. We observe that the LLM model LLaMA, finetuned with CoT explanation achieves better accuracy, J and BLEU score. We also notice that instruction-tuned generalized models attain almost perfect accuracy with very low BLEU score. We discuss the rationale in Section~\ref{sec:performance_instruction_llm}. Overall, our finetuned cross-platform models outperform the contemporary SoTA ParaDetox in terms of accuracy, J and BLEU score on the cross-platform dataset. We provide samples of models' responses in Appendix~\ref{appendix:modelsgeneration_sample}.
\LL{We now present a detailed discussion on the performance of the various models.}

\subsection{Comparison with SoTA}

We show the performance of the contemporary models (i.e., models with similar size to SoTA) in Table~\ref{table:result_cross_platform}. We find that both of our cross-platform finetuned BART and T5 outperform the SoTA ParaDetox on all metrics except BERTScore and Similarity. The better BERTScore and Similarity of ParaDetox can be attributed to its training dataset, which frequently transforms the toxic input with a minimal change (e.g., \LL{merely} deleting the strong words)~\citep{logacheva-etal-2022-paradetox}. It is to be noted that neither ParaDetox nor our models have seen data outside of Wikipedia, Reddit, and Twitter. However, our finetuned models still manage to exhibit superior performance compared to ParaDetox across the unseen platforms. We also find that DSS-based models outperform their respective explanation-based models in BLEU while lagging behind in accuracy. This is potentially because DSS is finetuned on detoxified output and explanation in a multitask setup. Although this helps the model align with the detoxified output separately (higher BLEU), it does not take explanation into account while detoxifying (hence, lower accuracy).

\subsection{Comparison to Instruction-Tuned LLMs}
\label{sec:performance_instruction_llm}

We compare our models' performance against the instruction-tuned LLMs. We notice that LLaMA-Chat, Alpaca, and Vicuna achieve perfect accuracy in some of the platforms. However, all of them lack in BLEU and BERTScore compared to the finetuned models. This is because they give priority to  generating non-toxic text over obeying input instructions that may involve toxic language. As a consequence, they often defy the instruction of detoxifying toxic inputs and frequently tend to produce generic statements such as: \textit{I'm sorry, but I cannot fulfill this request as it contains inappropriate language}. This incapability of detoxification by the generalized LLMs can potentially be attributed to the safety \LL{requirements} imposed during the pretraining and the consequent finetuning stages~\citep{llama2}. 
As a result, they receive high accuracy but very low BLEU score. Therefore, instruction-tuned models should not be deployed for the detoxification task without further finetuning, which also \LL{underscores}  the importance of training a dedicated instruction-tuned model for the detoxification task. We present a detailed discussion on the detoxification inability of the instruction-tuned LLMs in Appendix~\ref{appendix:inability_instruction_tuned_llms}.

\subsection{Improvement through Explanations}

As evident from Table~\ref{table:result_cross_platform}, CoT-expl LLaMA outperforms LLaMA-prompt and LLaMA-PD in terms of accuracy while the later two achieve better BERTScore. CoT explanation first helps the models identify the specific words or semantics that turns a text into toxic (see Appendix~\ref{appendix:models_explanation_sample} for  samples of models' generated explanation).
As a consequence, during the style transfer process, the models can focus on removing/modifying those specific portions to alleviate toxicity. Therefore, CoT-expl helps the models achieve better accuracy. However, identification of toxicity in an input text also means altering that input text. Hence, CoT-expl models achieve inferior BERTScore than vanilla models. Considering the nature of the detoxification task, it is more important to produce non-toxic text \LL{even if that causes} a few alterations to the input. Therefore, we prefer CoT-expl LLaMA model over the other models as the detoxification model of~\xdetox. 

\subsection{Performance on ParaDetox}

\begin{table}[ht]
\centering
\tiny
\resizebox{0.8\columnwidth}{!}{
\begin{tabular}{lcccccc}
\hline
\textbf{Model}                                                      & \textbf{Acc}   & \textbf{BS}    & \textbf{SIM} & \textbf{Fl} & \textbf{J}    & \textbf{BL}    \\ \hline
\cellcolor{red!15}\textbf{ParaDetox}                                                  & 90.16          & 96.65 & 85.63 & 88.52    &  68.34      & 69.99 \\
\cellcolor{orange!15}\textbf{T5-DSS}                                                     & 87.63          & 93.78  & 71.79        & 96.57   &  60.75       & 55.98          \\
\cellcolor{orange!15}\textbf{BART-DSS}                                                   & 92.10          & 93.68   &    67.41        & 96.27  &  59.77          & 52.38          \\
\cellcolor{mygreen!15}\textbf{T5-V}     & 91.21          & 93.81  & 70.57         & 95.23   &  61.23       & 54.78          \\
\cellcolor{mygreen!15}\textbf{T5-P}      & 89.42          & 93.97  &   71.98        & 94.93    &   61.10        & 55.47          \\
\cellcolor{mygreen!15}\textbf{T5-CE}    & 88.23          & 94.04   &   72.48         & 95.38  &  60.99        & 56.39          \\
\cellcolor{mygreen!15}\textbf{BART-V}   & 92.85          & 93.28  &  63.77        & 96.42    &   57.09      & 48.80          \\
\cellcolor{mygreen!15}\textbf{BART-P}    & 93.59          & 93.81   &   68.15         & 95.68   &   61.03       & 53.46          \\
\cellcolor{mygreen!15}\textbf{BART-CE}  & 93.29          & 93.01  &   63.02        & 96.72   &  56.86        & 48.74          \\ 
\cellcolor{yellow!15}\textbf{Alpaca}                                                     & 64.98	 & 94.36    &   80.74      & 96.72  &  54.59	& 54.23          \\
\cellcolor{yellow!15}\textbf{LLaMA-C}     & \textcolor{gray}{95.83}	& 88.80   &   56.84       & 97.76	&  52.43   & 23.29          \\
\cellcolor{yellow!15}\textbf{Vicuna}                                                     & 77.65	 & 90.43   &   69.13       & 97.91	&   54.05 & 29.63          \\ 
\cellcolor{red!15}\textbf{LLaMA-PD}  & 92.51 & \textbf{96.68} & \textbf{86.29} & 97.92	& \textbf{78.17} & \textbf{72.17} \\
\cellcolor{mygreen!15}\textbf{LLaMA-P}  & 93.89          & 92.72  &  60.72          & \textbf{98.06} & 55.09  &   42.55          \\
\cellcolor{mygreen!15}\textbf{LLaMA-CE} & \textbf{94.04} & 92.51  &  59.49        & 97.47   &   54.53       & 41.22          \\ \hline
\end{tabular}
}
\caption{
\label{table:result_paradetox}
Performance on the human annotated ParaDetox test set. 
Abbreviations are similar to Table~\ref{table:result_cross_platform}.
}
\end{table}

We further compare the models' performance against the human annotated parallel data. For this purpose, we evaluate the models on the test set of ParaDetox. As Table~\ref{table:result_paradetox} shows, we beat SoTA on accuracy and fluency. LLaMA-PD achieves the best similary, J, and BLEU score on this test set, which is unsurprising since this model has already been trained on this dataset. Notably, our suite of finetuned models still shows comparable BERTScore, while even outperforming LLaMA-PD and ParaDetox in terms of accuracy and fluency. This result indicates that although our dataset is artificially generated, the models trained on this dataset show impressive performance on human-annotated data, \LL{implying} the usability of our dataset.

\subsection{Paraphrase Detection}

We test the paraphrase detection capability of our finetuned BERT by passing a set of parallel detoxifiable and non-detoxifiable texts. For this purpose, we sample human-annotated parallel data (detoxifiable) from ParaDetox~\citep{logacheva-etal-2022-paradetox}. We also sample the human-labeled non-detoxifiable toxic data from ParaDetox and generate the corresponding non-toxic version with our finetuned detoxification model. Since the later set cannot be detoxified by humans, we consider these (toxic, non-toxic) pairs non-detoxifiable. We expect the paraphrase detection model to distinguish among detoxifiable and non-detoxifiable texts so that our framework can warn the users in case meaning is altered. We compare our model's performance against SoTA baselines finetuned on MRPC~\citep{mrpc} paraphrase detection task.

\begin{table}[h]
\centering
\resizebox{0.9\columnwidth}{!}{
\begin{tabular}{lcc}
\hline
\textbf{Model} & \textbf{Accuracy} & \textbf{F\textsubscript{1}-score} \\ \hline
BERT~\citep{bert}     & 79.33                & 80.88          \\
RoBERTa~\citep{roberta}           & 76.42                & 77.39          \\
ELECTRA~\citep{electra}        & 35.52                & 16.12          \\
TextAttack-BERT~\citep{textattack}        & 34.55                & 29.21          \\
TextAttack-RoBERTa~\citep{textattack}        & 28.96                & 13.61          \\
Sentence-BERT~\citep{sentence-bert}        & 50.00                & 66.63          \\
BERT \textbf{(ours)}    & \textbf{82.73}                & \textbf{83.13}          \\ \hline
\end{tabular}
}
\caption{
\label{table:performance_paraphrase}
Performance of the models on the paraphrase detection task. We compare our model's performance against SoTA baselines finetuned on MRPC~\citep{mrpc} dataset. \textbf{Bold} font represents the best performance for a particular metric.
}
\end{table}

\noindent We present results in Table~\ref{table:performance_paraphrase}. As evident, our paraphrase detector \LL{comfortably} outperforms the SoTA baselines. This shows the importance of a dedicated paraphrase detector in our framework, since models trained on generic paraphrase datasets may fail to transfer their knowledge when comparing the semantic meaning between toxic/non-toxic pairs.

\section{Analyses}

\subsection{Effectiveness of Cross-Platform Data}

We further analyze how our cross-platform dataset improves models trained on human-annotated data. Hence, we take the finetuned ParaDetox model and continue training it on our cross-platform dataset with varying sample sizes ($100$ - $1000$ samples). Then, we evaluate the models' performance on the human-annotated ParaDetox test set.

\begin{figure}[h]
    \centering
  \includegraphics[width=0.7\columnwidth]{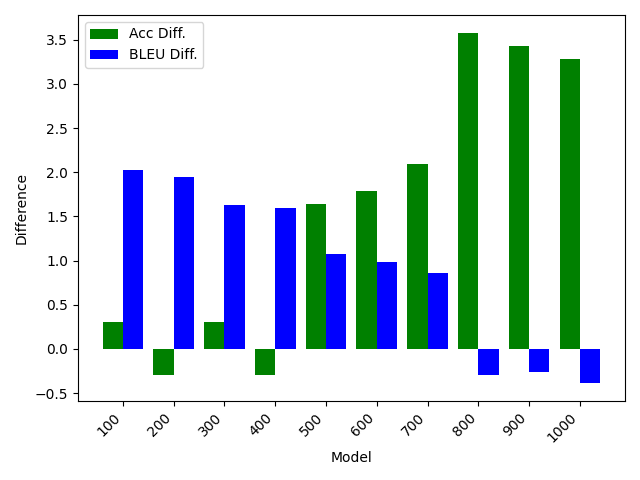}
  \caption{\label{fig:paradetox_sample}
  Difference in accuracy and BLEU between the finetuned Paradetox and the original ParaDetox. 
  }
\end{figure}

Figure~\ref{fig:paradetox_sample} shows the relative difference in accuracy and BLEU between the ParaDetox model trained on different sample sizes of our cross-platform dataset and the original ParaDetox model. As is evident, the finetuned models (up to sample size $700$) tend to maintain higher BLEU score. Importantly, the model's accuracy tends to increase with the increase in the sample size. The higher accuracy and BLEU score signify the models' capability to detoxify input text while producing human-like non-toxic output, which consequently indicates the effectiveness of our cross-platform dataset. We report the detailed results in Appendix~\ref{appendix:paradetox_sample}. We further present analysis on multilingual transfer of detoxification in Appendix~\ref{sec:multilingual}.

\subsection{Performance on Implicit Hate Speech}
\label{sec:implicit_hate_speech}

To analyze the models' behavior on implicit and adversarial hate speech datasets, we apply the models on ToxiGen~\citep{hartvigsen-etal-2022-toxigen}, a machine-generated dataset containing implicit and adversarial hate speech. For the detoxification task, we select the human-annotated samples from the test set with toxicity ratings over $3$ out of $5$. We first generate a non-toxic version of this test set with the detoxification models, then compute BERTScore as well as the non-toxic accuracy of the models using \texttt{Toxicity\_RoBERTa}~\citep{logacheva-etal-2022-paradetox} and \texttt{ToxiGen\_RoBERTa}~\citep{hartvigsen-etal-2022-toxigen}. 

\begin{figure}[h]
  \centering
  \includegraphics[width=0.8\columnwidth]{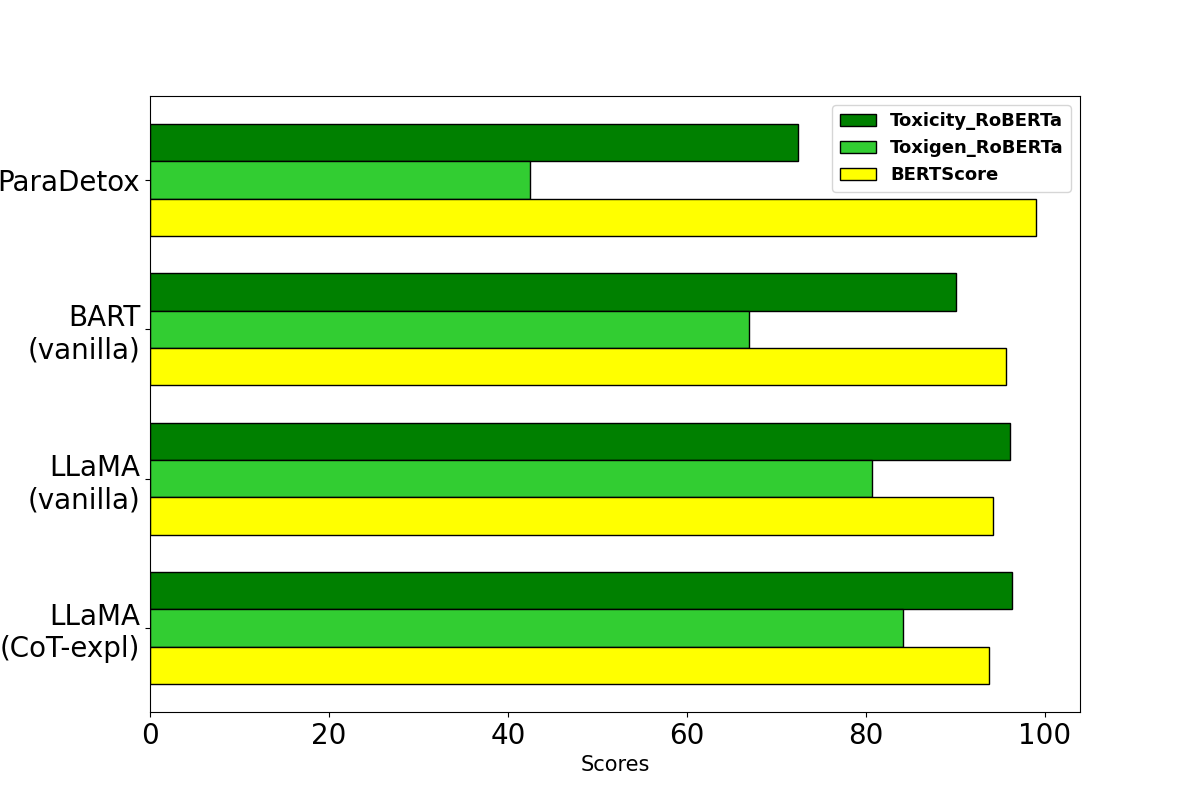}
  \caption{\label{fig:performance_toxigen}
  \texttt{Toxicity\_RoBERTa} (accuracy), \texttt{ToxiGen\_RoBERTa} (accuracy), and BERTScore of the models on ToxiGen test set. 
  }
\end{figure}

As Figure~\ref{fig:performance_toxigen} shows, our  models produce less toxicity compared to the SoTA ParaDetox. Specifically, our finetuned BART performs better than ParaDetox, while CoT-expl LLaMA performs the best in terms of  accuracy while maintaining an impressive BERTScore. The high accuracy of our models on this implicit toxicity dataset signifies that~\xdetox~is more capable of countering implicit hate speech \LL{than merely} depending on searching and removing explicit toxic words.

\subsection{Robustness of~\xdetox}
\label{sec:token_adversaries}

\textbf{Curated token-level adversaries.} Due to  censorship reasons, users tend to mask out specific portion of a strong word (e.g., `\textit{f\#ck}', `\textit{sh*t}', etc) while commenting on social platforms. Although these masked words are still understandable from a human perspective, how the models perceive these words is unclear. To study the models' abilities to detect adversarial strong tokens, we carefully curate a list of $15$ texts containing different levels of masked words. We pass them to the models to generate non-toxic versions and then manually inspect the outputs.

\begin{table}[h]
\centering
\resizebox{0.7\columnwidth}{!}{
\begin{tabular}{ccc}
\hline
\textbf{Models}  & \texttt{\textbf{Toxicity}} & \texttt{\textbf{ToxiGen}}\\ \hline
ParaDetox        & 93.32	 & 84.88                                                              \\  \hline
BART-V \textit{(ours)}           &  96.86	&  95.1                                                             \\ \hline
LLaMA-CE \textit{(ours)}       & \textbf{97.21}	&  \textbf{96.22}                                                              \\ \hline
\end{tabular}
}
\caption{
\label{tab:token_adversaries}
Performance of the models on the automated token-level adversaries. $2$\textsuperscript{nd} and $3$\textsuperscript{rd} columns represent the non-toxic performance using \texttt{Toxicity\_RoBERTa} and \texttt{ToxiGen\_RoBERTa} classifiers respectively.
}
\end{table}



We find that ParaDetox, our BART-V, and our LLaMA-CE produce two, eight, and $12$ non-toxic and meaning-preserving responses, respectively (see Appendix~\ref{appendix:token_level_adversaries}). We further notice that~\xdetox~ (LLaMA-CE) is more successful in identifying adversarial words and as a result produces non-toxic versions of the toxic texts. 

\noindent \textbf{Large-scale, automated adversaries.} We additionally conduct a large-scale analysis on a generated list of $5,000$ sentences (see Appendix~\ref{appendix:automated_token_level_adversaries} for details). We then calculate model accuracy using \texttt{Toxicity\_RoBERTa} and \texttt{ToxiGen\_RoBERTa}. Table~\ref{tab:token_adversaries} shows LLaMA-CE exhibits the highest accuracy \eat{by both models} followed by BART-V. This further substantiates the usefulness of our dataset as well as the detoxification models finetuned on this dataset in the identification of adversarial toxic words.

\section{Human Evaluation}
\label{sec:human_eval}



\noindent\textbf{Evaluation Setup.} Following~\citet{self_instruct_wang,wu2023laminilm,khondaker-etal-2023-gptaraeval}, we implement a four-level (A, B, C, D) rating system to measure the detoxification responses from the model. To handle non-detoxifiability, we incorporate two additional ratings, namely, N (\underline{n}on-toxic) and T (\underline{t}oxic or generic statements). We randomly sample $200$ texts from our cross-platform dataset and ask two pairs of fluent English speakers (total = $4$) to rate models' responses (see Appendix~\ref{appendix:human_evaluation} for details). 

\begin{figure}[h]
     \centering
     \begin{subfigure}[t]{0.5\textwidth}
         \centering
         \includegraphics[width=\textwidth]{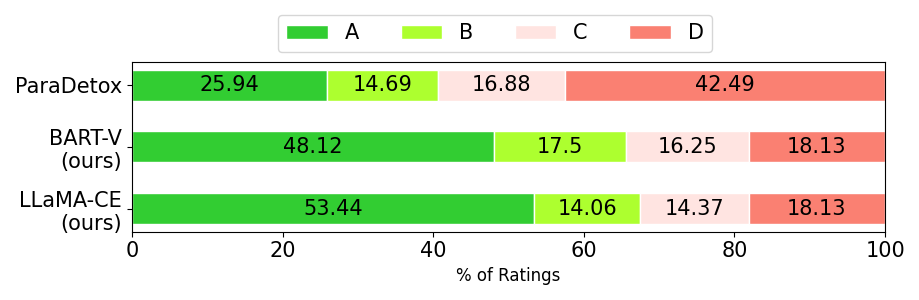}
           \caption{Human evaluation on detoxifiable inputs.}
        \label{fig:human_eval_rating_detoxifiable}
     \end{subfigure}
    \hfill
     \begin{subfigure}[t]{0.5\textwidth}
         \centering
         \includegraphics[width=\textwidth]{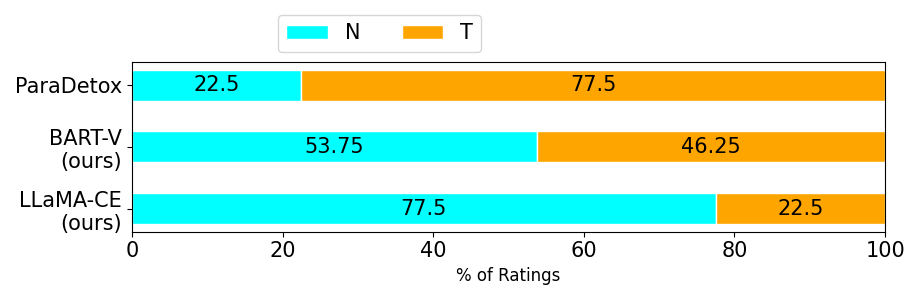}
         \caption{Human evaluation on non-detoxifiable inputs.}
        \label{fig:human_eval_rating_nondetoxifiable}
     \end{subfigure}
        \caption{\label{fig:human_eval_rating}
        Human evaluation on the models' responses. \textcolor{mygreen}{A} is the best, and \textcolor{red}{D} is the worst rating for detoxifiable input. \textcolor{cyan}{N} is the good and  \textcolor{orange}{T} is the bad rating for non-detoxifiable input.}
\end{figure}

\noindent\textbf{Results.} We report the results in Figure~\ref{fig:human_eval_rating} (inter-annotator agreement $=$ $0.67$). We find that detoxification responses produced by \xdetox~(LLaMA-CE) and BART-V are rated as mostly of fine quality. Specifically, our \xdetox~($67.50$\%) and BART-V ($65.62$\%) \LL{provide} more non-toxic and meaning preserving-responses (ratings \textit{A} and \textit{B}) compared to the SoTA ParaDetox model ($40.63$\%). For non-detoxifiable input, \xdetox~exhibits more robustness with $55$\% less toxic output than ParaDetox.


\section{Human Evaluation of Explanation}
\label{sec:main_human_expl}

\modification{To assess the quality of the toxicity explanation, we conduct another human evaluation similar to the detoxification evaluation. We implement a four-level (A, B, C, D) rating system to measure the quality of the explanation generated by the models. We randomly sample $100$ test cases and pass it to two human annotators for evaluating the explanation. We assess the quality of the explanations based on the following metrics:}
\begin{itemize}
    \item \textbf{Relevance:} How relevant is the explanation given the context of the toxic input?
    \item \textbf{Comprehensiveness:} How comprehensive is the explanation? E.g., Can the model correctly identify the toxic terms in the input?
    \item \textbf{Convincing:} How persuasive is the explanation? In other words, will the user be convinced enough regarding the toxicity of the input text so that they will agree to alter it?
\end{itemize}

We provide a detailed description of the evaluation framework in Appendix~\ref{appendix:explanation_framework}.

\begin{figure}[h]
     \centering
     \includegraphics[width=\columnwidth]{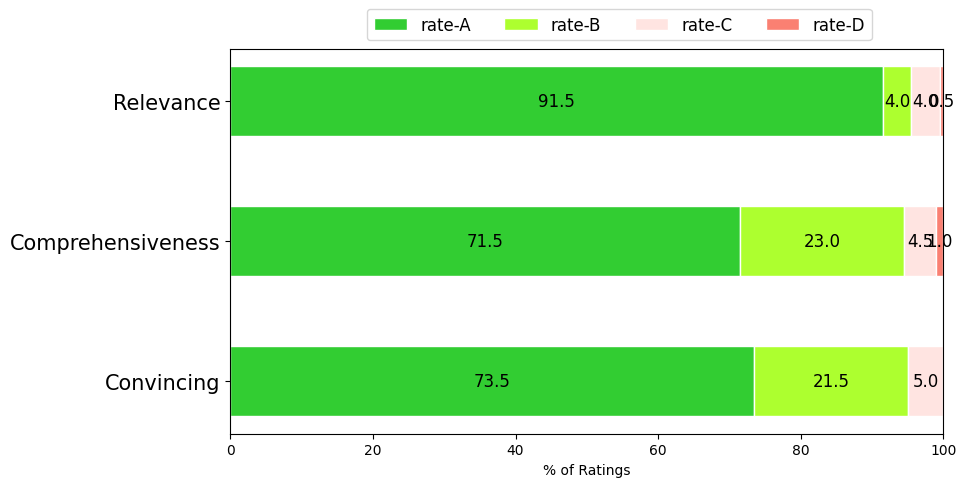}

    \caption{\label{fig:human_expl_rating_chatgpt}
    Human evaluation on the explanations (generated by \texttt{ChatGPT}) for the toxic inputs from training dataset (inter-annotator agreement $=$ $0.78$). \textcolor{mygreen}{A} is the best, and \textcolor{red}{D} is the worst rating for explanation of the toxic input.}
\end{figure}

\begin{figure}[h]
     \centering
     \begin{subfigure}[t]{0.45\textwidth}
         \centering
         \includegraphics[width=\textwidth]{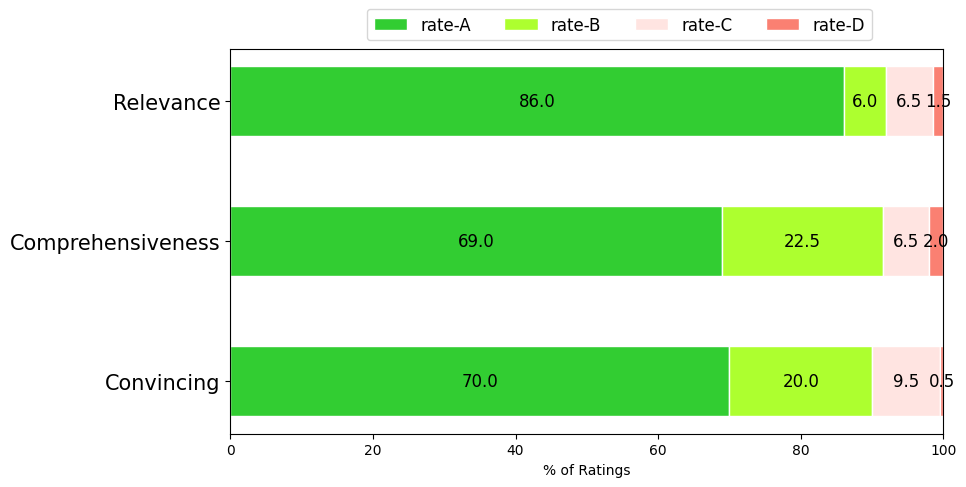}
           \caption{Human evaluation on \xdetox~(LLaMA-CE) generated explanation for the toxic input.}
        \label{fig:human_expl_rating_llama}
     \end{subfigure}
    \hfill
     \begin{subfigure}[t]{0.45\textwidth}
         \centering
         \includegraphics[width=\textwidth]{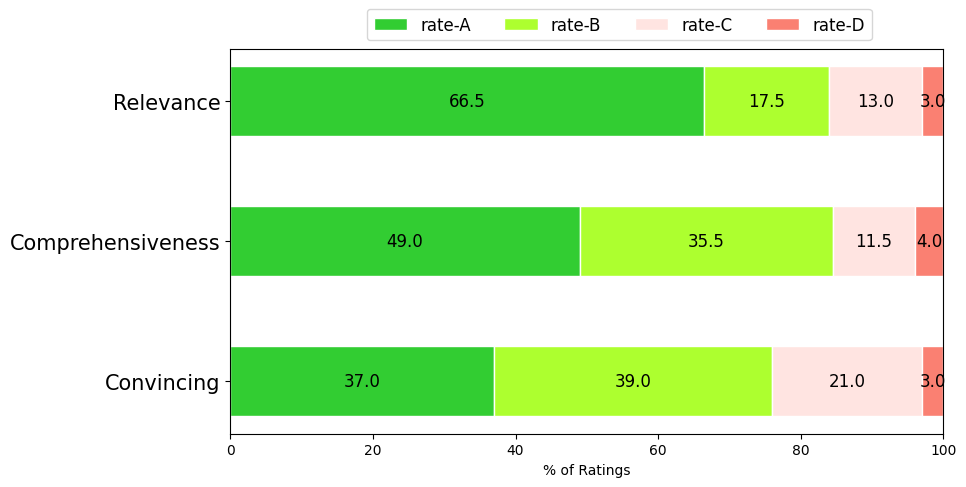}
         \caption{Human evaluation on BART generated explanation for the toxic input.}
        \label{fig:human_expl_rating_bart}
     \end{subfigure}
        \caption{\label{fig:human_expl_rating}
        Human evaluation of the models' generated explanation for the toxic inputs (inter-annotator agreement $=$ $0.65$). \textcolor{mygreen}{A} is the best, and \textcolor{red}{D} is the worst rating for explanation of the toxic input.}
\end{figure}

\noindent \modification{\textbf{Quality of training data.} We first analyze the quality of the training data (explanation) generated by \texttt{ChatGPT} (Figure~\ref{fig:human_expl_rating_chatgpt}). Through human evalution, we find that \texttt{ChatGPT} produces mostly relevant, comprehensive, and convincing explanations. This human evaluation further demonstrates the high quality of our training data.} 

\noindent \modification{\textbf{Results.} We present the evaluation results in Figure~\ref{fig:human_expl_rating}. As noticed, \xdetox~(LLaMA-CE) (Figure~\ref{fig:human_expl_rating_llama}) produces better explanations according to the human annotators. We further find that the majority of the explanations from \xdetox~(Figure~\ref{fig:human_expl_rating_llama}) are relevant ($86$\% of rating-A) and comprehensive ($69$\% of rating A). Importantly, $70$\% (rating-A) of the responses from \xdetox~are found convincing, signifying that the user would be motivated enough to alter the input.}

\section{Conclusion}
In this work, we propose \xdetox, a comprehensive end-to-end detoxification framework to tackle toxic language across multiple platforms. We generate a novel cross-platform pseudo-parallel corpus through multi-step data processing and generation with \texttt{ChatGPT}. We train a suite of detoxification models. Especially, our corss-platform detoxification model trained with CoT explanation (CoT-expl LLaMA) outperforms SoTA detoxification models. We additionally introduce explanation into the \xdetox~framework for promoting  trustworthiness. We also develop a dedicated paraphrase detector to handle the cases of non-detoxifiability. Through an extensive experimental analysis, we further show the effectiveness of our cross-platform data as well as the robustness of \xdetox~against implicit and token-level adversarial toxicity. 


\section{Limitations and Ethics Statement}
\subsection{Limitations}

\noindent \textbf{Data Generation Process.} In this work, we use \texttt{ChatGPT}, a gpt-3.5-turbo version from June, 2023. Since the model can be updated on a regular interval, the prompting strategy and the data generation pipeline discussed in Section~\ref{sec:Proposed_Methodology} should be treated accordingly, since the model's responses can change over time~\citep{chen2023chatgptchange}.

\noindent \textbf{Data Quality.} We propose an automated data generation pipeline to create a pseudo-parallel cross-platform corpus (\S\ref{sec:Proposed_Methodology}). Our synthetic data generation process involves multi-stage data processing without the necessity of direct human inspection. Although this automated pipeline makes the overall data generation process scalable, it comes at the risk of allowing low-quality data in our cross-platform corpus. Hence, we suggest human inspection to remove any sort of potential vulnerability and maintain a standard quality of the corpus. Additionally, we combine datasets from multiple platforms. Since the toxicity nature of language is often deemed as subjective~\citep{sap,wilds}, the level of toxicity may vary across the platforms based on the context.

\noindent \textbf{Model Responses.} We show that \xdetox~exhibits impressive ability in generating detoxified responses. However, looking at the results (\S\ref{sec:Results}), we believe there is still room for improvement for the models in terms of producing meaning-preserved detoxified outcomes. Moreover, as evident from our analyses in Section~\ref{sec:implicit_hate_speech} and Section~\ref{sec:token_adversaries}, models can be vulnerable to implicit, adversarial tokens and continue to produce toxic content. Therefore, we recommend that \xdetox~should be couched with caution before deployment. 

\noindent \textbf{Model Evaluation.} We use six automated metrics (Accuracy, BERTScore, Content Similarity, Fluency, J, and BLEU) to evaluate our models. As noticeable from Section~\ref{sec:Results}, depending on a single metric to measure the models' performance can be deceptive. Since detoxification is a form of style transfer task and there is still a lack of an effective method for aggregating the aforementioned metrics~\citep{ostheimer-eval-2023}, we suggest not depending on a particular metric and looking at the performance of models holistically.

\noindent \textbf{Findings.} Some of our findings suggest that instruction-tuned LLMs often deny following instructions while dealing with toxic input (\S\ref{sec:performance_instruction_llm}) and produce a generic statement. We hypothesize it may be the case because of the safety measurement imposed on these models. This scenario can occur for some particular tasks like detoxification that require handling toxic inputs. However, we believe that further instruction-tuning of these models on tasks like detoxification can alleviate the problem.

\subsection{Ethics Statement}
\noindent \textbf{Data Collection and Release.} As mentioned in Section~\ref{sec:data_collection}, we compile datasets from a wide range of platforms. The sources of the datasets are primarily collected from~\citet{unified_datasets} and~\citet{hatespeech_datasets}. To ensure proper credit assignment, we refer users to the original publications in Table~\ref{table:datasets}. We create the cross-platform detoxification corpus for academic research purposes. We would also like to mention that some content of Figure~\ref{fig:xdetox_intro} and Figure~\ref{fig:methodology} are generated using \texttt{GPT-4} and \texttt{DALL-E} for illustration purposes.

\noindent \textbf{Intended Use.} The intended use of \xdetox~is for the detoxification tasks. We aim to help researchers to build an end-to-end complete detoxification framework. \xdetox~can also be regarded as a promising baseline to develop more robust and effective detoxification frameworks.

\noindent \textbf{Potential Misuse and Bias.}
Our detoxification corpus and models can potentially be misused to generate toxic and biased content. For these reasons, we recommend that \xdetox~not be used in applications without careful prior consideration of potential misuse and bias.

\section*{Acknowledgments}\label{sec:acknow}
We acknowledge support from Canada Research Chairs (CRC), the Natural Sciences and Engineering Research Council of Canada (NSERC; RGPIN-2018-04267, RGPIN-2020-05408), the Social Sciences and Humanities Research Council of Canada (SSHRC; 895-2020-1004; 895-2021-1008), Canadian Foundation for Innovation (CFI; 37771), Digital Research Alliance of Canada,\footnote{\href{https://alliancecan.ca}{https://alliancecan.ca}} and UBC Advanced Research Computing-Sockeye.\footnote{\href{https://arc.ubc.ca/ubc-arc-sockeye}{https://arc.ubc.ca/ubc-arc-sockeye}}

\bibliography{anthology,custom,bib_scl_fish}

\appendix

\clearpage
\appendixpage
\addappheadtotoc
\counterwithin{figure}{section}
\counterwithin{table}{section}

\section{Related Works}
\label{appendix:related_works}

\subsection{Abusive Language Detection}

 Over the years, the task of abusive language detection has been studied in NLP in the form of hate speech~\cite{founta,davidson,golbeck}, sexism/racism~\cite{waseem}, cyberbulling~\cite{xu,dadvar}.
Earlier works in abusive language detection depend on feature-based approaches to identify the lexical difference between abusive and non-abusive language~\cite{warner,waseem,ribeiro}.
Recently, Transformer-based~\cite{transformer} architectures like BERT~\cite{bert}, RoBERTa~\cite{roberta} have been introduced in the abusive language detection task~\cite{nuli,swamy}. However, the study of~\citet{hatespeech_datasets} raises the concern that most of the prior works on abusive language detection focus on a single platform due to the inaccessibility to multiple platforms and thus, do not scale well on other platforms~\citet{schmidt}. As a result,~\citet{karan,grondahl} point out that the models are not suitable to apply to other platforms due to the lack of generalization. To alleviate this issue,~\citet{khondaker-2023-scl-fish} more recently propose a meta-learning algorithm to detect abusive language across different platforms.

\subsection{Text Style Transfer}

Text style transfer (TST) refers to rephrasing the style of a source text (e.g. sentiment, politeness) into a target style while changing the meaning of the input as little as possible~\citep{reid-zhong-2021-lewis}. TST has been explored in the field of NLP due to its applications in sentiment transfer~\citep{shen-style-2017}, formality transfer~\citep{rao-formality-2018}, authorship attribute transfer~\citep{shetty-authorship-2017,patel-authorship-2022}, or increasing politeness~\citep{niu-politeness-2018,madaan-politeness-2020,mukherjee-politeness-2023}. The lack of parallel datasets is one of the main bottlenecks for TST tasks~\citep{liu-lamer-2021}. To alleviate this challenge, several unsupervised methods have been proposed~\citep{zhang-unsupervised-2020,liu-lamer-2021}. Specifically,~\citet{liu-data_aug-2022} create a pseudo-parallel corpus via data augmentation to circumvent the lack of human-annotated parallel data. Prior works~\citep{gong-style-2019,wang-style-2019} also resort to an auxiliary classifier to guide the style of the generated text. With the advancement of large language models (LLMs), recent works~\citep{patel-llm-2022,pu-llm-2023} employ LLMs like GPT-3~\citep{gpt3} for parallel data generation and style transfer tasks. Studies like \citet{reif-llm-2022} show the effectiveness of LLMs in TST, while~\citet{steer-choi-2023} remove the cost of human supervision by creating a synthetic pseudo-parallel style transfer dataset with reinforcement learning. 

\section{Experimental Details}
\label{appendix:experimental_details}

\subsection{Models Comparison}

\noindent \colorbox{red!15}{\textbf{SoTA Baseline.}} We compare our models with the state-of-the-art detoxification model, \textbf{ParaDetox}~\citep{logacheva-etal-2022-paradetox}; a BART-based model finetuned on crowdsourced parallel detoxification corpus. The model is trained on three platforms, namely, Jigsaw~\citep{jigsaw} (Wikipedia's talk edit page), Reddit, and Twitter (now known as X). We evaluate this model without further finetuning on our dataset to determine the efficacy of the model on the cross-platform detoxification task. For fair comparison with our cross platform models, we also finetune a LLaMA~\citep{llama2} model on the ParaDetox training set.



\noindent \colorbox{orange!15}{\textbf{*-DSS.}} We additionally compare our models with SoTA distillation method, \textit{Distilling Step-by-Step} (DSS) proposed by~\citet{dss-Hsieh}. We use DSS method to distill both detoxification and explanation coming from \texttt{ChatGPT} into BART and T5 models. Following the work, we use a multitask framework to combine the training of generating both non-toxic version and explanation from the models given a toxic input.

\noindent \colorbox{yellow!15}{\textbf{Instruction-tuned.}} We evaluate the performance of generic instruction-tuned models like Alpaca~\citep{alpaca}, instruction-tuned LLaMA (Chat)~\citep{llama2}, and Vicuna~\citep{vicuna} on the cross-platform detoxification tasks. We use the corresponding 7B versions for all the models. These models are already finetuned on a wide range of generic tasks. Hence, we omit these models from further finetuning on our cross-platform dataset to examine the generalizability of these models.

\noindent \colorbox{mygreen!15}{\textbf{Cross-Platform Models.}} We finetune a suit of models on the cross-platform datasets. In particular, we finetune BART and T5 to directly compare against the contemporary SoTA (e.g., ParaDetox). We further finetune LLM like LLaMA to observe the performance of LLM as well as compare against generic instruction-tuned models (e.g., Alpaca). As discussed in Section~\ref{sec:model_training}, we finetune our models in multiple setups. For T5 and BART, we (1) direct finetune the model to generate non-toxic version given the toxic version \underline{(vanilla)}; (2) concatenate a prompt with toxic version as the model input \underline{(prompt)}; (3) employ CoT finetuning to instruct the model explain why the given input is toxic before generating the non-toxic version \underline{(CoT-expl)}. For LLaMA finetuning, we use the two variations mentioned above namely, (1) prompt and (2) CoT-expl.

\subsection{Performance Metrics}

We report the models' performance on seven unseen platforms (Table~\ref{table:result_cross_platform}) as well as the overall average performance across the platforms. We evaluate the models based on the following metrics.

\noindent \textbf{Accuracy.} Following~\citet{logacheva-etal-2022-paradetox}, we compute the accuracy of the models based on the percentage of non-toxic outputs identified by a style classifier. We use the same RoBERTa style classifier as the authors.

\noindent \textbf{BERTScore.} We use BERTScore to compute how the models preserve the semantic meaning. Specifically, we utilize SimCSE~\citep{simcse} RoBERTa-large model to obtain the embeddings of the input-output pair and then measure the similarity between them.

\modification{\noindent \textbf{Content Similarity.} Cosine similarity between the embeddings of the original text and the output computed with the model of~\citet{wieting-etal-2019-beyond}. This model is trained on paraphrase pairs extracted from ParaNMT corpus.}

\noindent \textbf{Fluency.} Following~\citet{logacheva-etal-2022-paradetox}, we measure the percentage of fluent sentences identified by a RoBERTa-based classifier trained on the linguistic acceptability (CoLA) dataset~\citep{cola-warstadt}.

\modification{\noindent \textbf{Joint Metric.} An aggregated metric of the multiplication of three imdividual metrics \textit{Accuracy}, \textit{Content Similarity}, and \textit{Fluency} proposed by~\citet{logacheva-etal-2022-paradetox}.}

\noindent \textbf{BLEU.} We compute the BLEU score between the generated non-toxic version and the original non-toxic version on the test set.

\subsection{Implementation Details}
\label{appendix:implementation_details}

For finetuning cross-platform detoxification models, we use pretrained models (T5-base, BART-base, and LLaMA-2-7b) from \textit{Huggingface}~\citep{wolf_huggingface}. We set the maximum source length of $128$ tokens for T5, BART and $512$ tokens for LLaMA. We set the maximum target length to $256$ with explanation and $128$ without explanation for T5, BART. On the other hand, we use the maximum target length of $512$ for LLaMA for both cases. We use a batch size of $32$ for T5, BART and a batch size of $8$ with the gradient accumulation step of $8$ for LLaMA. For all the models, we set the learning rate to $3$e-$5$ with \textit{cosine} scheduler and a warmup ratio of $0.03$. We train T5 and BART for $15$ epochs while LLaMA for $10$ epochs and choose the best respective models based on the validation set performance. We use $1$ Nvidia A100 40GB GPU to train T5, BART and $4$ Nvidia A100 40GB GPUs to train LLaMA.

For finetuning the paraphrase detection model, We use the pretrained BERT-base (uncased) from \textit{Huggingface}~\citep{wolf_huggingface} as the backbone architectures. We set the maximum sequence length to $128$ for both toxic and non-toxic input pairs. We use a batch size of $32$ and a learning rate of $5$e-$5$. We train the models for $50$ epochs and select the best models based on the models' validation set performance.

\section{Performance on Other Platforms}
\label{appendix:performance_other_cross_platforms}

\begin{table*}[t]
\centering
\resizebox{\linewidth}{!}{
\begin{tabular}{l|cccccc|cccccc|cccccc|cccccc|}
\hline
                   & \multicolumn{6}{c|}{\cellcolor{pink!20}\textbf{convai}}                                                        & \multicolumn{6}{c|}{\cellcolor{yellow!20}\textbf{gab}}                                                           & \multicolumn{6}{c|}{\cellcolor{magenta!20}\textbf{hatecheck}}                                                     & \multicolumn{6}{c|}{\cellcolor{cyan!10}\textbf{stormfront}}                                                    \\ \hline
\textbf{Model}     & \textbf{ACC} & \textbf{BS} & \textbf{SIM} & \textbf{FL} & \textbf{J} & \textbf{BL} & \textbf{ACC} & \textbf{BS} & \textbf{SIM} & \textbf{FL} & \textbf{J} & \textbf{BL} & \textbf{ACC} & \textbf{BS} & \textbf{SIM} & \textbf{FL} & \textbf{J} & \textbf{BL} & \textbf{ACC} & \textbf{BS} & \textbf{SIM} & \textbf{FL} & \textbf{J} & \textbf{BL} \\ \hline
\cellcolor{red!15}\textbf{ParaDetox}          & 82.00        & 95.93       & 75.71        & 97.00       & \textbf{60.22}      & 31.66       & 80.00        & 94.49       & 76.79        & 83.00       & 50.99      & 25.41       & 24.00        & \textbf{98.02}       & \textbf{87.68}        & 98.00       & 20.62      & 22.68       & 88.00        & \textbf{96.83}       & \textbf{81.80}        & 95.00       & 68.38      & 40.38       \\
\cellcolor{orange!15}\textbf{T5-DSS}             & 64.84        & 95.65       & 76.09        & 94.51       & 46.63      & 38.91       & 68.89        & 95.41       & 77.67        & 96.67       & 51.73      & 40.39       & 68.09        & 95.20       & 73.75        & 95.74       & 48.08      & 38.00       & 63.04        & 96.11       & 80.27        & 98.91       & 50.05      & 38.89       \\
\cellcolor{orange!15}\textbf{BART-DSS}          & 84.62        & 93.84       & 67.00        & 98.90       & 56.07      & 42.51       & 81.11        & 93.79       & 69.56        & 98.89       & \textbf{55.79}      & 47.26       & 82.98        & 93.81       & 68.01        & \textbf{100.00}      & 56.43      & 40.96       & 88.04        & 94.06       & 71.79        & 98.91       & 62.51      & 43.55       \\
\cellcolor{mygreen!15}\textbf{T5-V}        & 81.00        & 93.71       & 65.91        & \textbf{100.00}      & 53.39      & 36.76       & 82.00        & 93.65       & 68.09        & 97.00       & 54.16      & 38.52       & 50.00        & 94.17       & 57.82        & \textbf{100.00}      & 28.91      & 31.87       & 85.00        & 93.50       & 63.76        & 99.00       & 53.65      & 40.18       \\
\cellcolor{mygreen!15}\textbf{T5-P}          & 80.00        & 91.71       & 54.46        & 99.00       & 43.13      & 35.93       & 78.00        & 91.74       & 56.93        & 96.00       & 42.63      & 38.89       & 46.00        & 90.50       & 34.29        & 99.00       & 15.62      & 30.42       & 87.00        & 90.50       & 49.43        & 99.00       & 42.57      & 39.91       \\
\cellcolor{mygreen!15}\textbf{T5-CE}       & 68.13        & 89.41       & 40.65        & 94.51       & 26.17      & 33.13       & 76.67        & 89.74       & 42.01        & 97.78       & 31.49      & 38.23       & 82.98        & 89.37       & 40.31        & 94.68       & 31.67      & 34.91       & 72.83        & 90.12       & 43.46        & 96.74       & 30.62      & 33.58       \\
\cellcolor{mygreen!15}\textbf{BART-V}      & 89.00        & 93.00       & 60.50        & 99.00       & 53.31      & 34.11       & 90.00        & 92.55       & 65.34        & 93.00       & 54.69      & 41.06       & 73.00        & 93.58       & 52.30        & 99.00       & 37.80      & 34.31       & 92.00        & 94.24       & 69.35        & 99.00       & 63.16      & 45.43       \\
\cellcolor{mygreen!15}\textbf{BART-P}       & 85.00        & 91.34       & 51.94        & 98.00       & 43.27      & 35.49       & 87.00        & 90.78       & 53.33        & 96.00       & 44.54      & 40.10       & 67.00        & 89.73       & 28.61        & 98.00       & 18.79      & 32.51       & 87.00        & 91.73       & 51.26        & 96.00       & 42.81      & 44.20       \\
\cellcolor{mygreen!15}\textbf{BART-CE}     & 89.01        & 89.24       & 37.98        & 97.80       & 33.06      & 35.62       & 85.56        & 89.30       & 38.86        & 98.89       & 32.88      & 40.96       & 90.43        & 89.01       & 36.65        & 98.94       & 32.79      & 34.63       & 86.96        & 89.54       & 41.08        & 97.83       & 34.95      & 34.55       \\
\cellcolor{yellow!15}\textbf{Alpaca}    & 45.05        & 84.81       & 17.05        & 97.80       & 7.51       & 9.42        & 50.00        & 84.31       & 21.49        & 93.33       & 10.03      & 9.12        & 46.81        & 86.13       & 9.96         & 96.81       & 4.51       & 7.65        & 48.91        & 84.53       & 16.55        & 96.74       & 7.83       & 7.29        \\
\cellcolor{yellow!15}\textbf{LLaMA-C} & \textcolor{gray}{98.90}        & 84.46       & 19.10        & 98.90       & 18.68      & 10.34       & \textcolor{gray}{97.78}        & 83.84       & 25.52        & 98.89       & 24.68      & 14.04       & \textcolor{gray}{97.87}        & 85.39       & 9.34         & 98.94       & 9.04       & 7.81        & \textcolor{gray}{97.83}        & 84.17       & 17.25        & 98.91       & 16.69      & 7.40        \\
\cellcolor{yellow!15}\textbf{Vicuna}    & 81.32        & 84.68       & 17.63        & 97.80       & 14.02      & 10.10       & 80.00        & 84.29       & 21.77        & 96.67       & 16.84      & 11.47       & 84.04        & 85.81       & 9.38         & 98.94       & 7.80       & 9.10        & 84.78        & 84.43       & 18.45        & \textcolor{gray}{100.00}      & 15.64      & 8.41        \\
\cellcolor{red!15}\textbf{LLaMA-PD}     & 83.31	& \textbf{96.22}	& \textbf{77.32}	& 97.59	& 62.86	& 35.28	& 84.56	& \textbf{95.45}	& \textbf{78.24}	& 98.92	& 65.45	& 32.42	& 32.58	& 96.62	& 86.34	& 98.58	& 27.73	& 29.53	& 89.37	& 96.52	& 81.46	& 99.00	& \textbf{72.07}	& 44.63      \\
\cellcolor{mygreen!15}\textbf{LLaMA-P}     & 87.91        & 92.04       & 55.08        & 96.70       & 46.82      & 58.25       & 93.33        & 91.89       & 57.01        & \textbf{100.00}      & 53.21      & \textbf{61.47}       & 93.62        & 91.49       & 57.09        & 98.94       & 52.88      & \textbf{57.05}       & \textbf{94.57}        & 92.25       & 58.67        & \textbf{100.00}      & 55.48      & 59.60       \\
\cellcolor{mygreen!15}\textbf{LLaMa-CE}    & \textbf{92.31}        & 88.80       & 57.36        & 97.80       & 51.78      & \textbf{59.96}       & \textbf{97.78}        & 91.67       & 57.25        & 98.89       & 55.36      & 61.27       & \textbf{95.74}        & 89.46       & 57.12        & 98.94       & \textbf{54.11}      & 54.86       & 93.48        & 90.14       & 59.54        & 98.91       & 55.05      & \textbf{60.25}       \\ \hline
\end{tabular}
}
\caption{
\label{table:result_rest_cross_platform}
Performance of the models on the rest of the cross-platform datasets. \textbf{Acc} = percentage of non- toxic outputs identified by a style classifier, \textbf{BS} = BERTScore, \textbf{Sim} = Content Similarity, \textbf{Fl} = Fluency, \textbf{J} = Joint Metric, \textbf{BL} = BLEU Score. \textbf{V} = Vanilla, \textbf{P} = Prompt, \textbf{PD} = ParaDetox-finetuned, \textbf{CE} = CoT-expl, \textbf{C} = Chat. \textbf{Bold} font represents the best performance for a particular metric. We separately show the best performance of the \colorbox{yellow!15}{instruction-tuned models} in \textcolor{gray}{gray} due to their inability to detoxification (Section~\ref{sec:performance_instruction_llm}).
}
\end{table*}

\modification{We provide the models' performances on the rest of the platform in Table~\ref{table:result_rest_cross_platform}.}

\section{\xdetox~Across Platforms}
\label{appendix:accuracy_cross_platforms}

We evaluate the accuracy of non-toxicity generated by the models using the corresponding in-platform classifiers. For this purpose, we use the six in-house classifiers (Section~\ref{sec:data_filtration}) to compute the accuracy of their respective datasets. 

\begin{table}[h]
\centering
\resizebox{\columnwidth}{!}{
\begin{tabular}{lccccccc}
\hline
\textbf{Model}                                                      & \textbf{Overall}        & \textbf{wikipedia} & \textbf{twitter}        & \textbf{fb\_yt}         & \textbf{HateCheck}      & \textbf{stormfront}     & \textbf{convAI}         \\ \hline
\cellcolor{red!15}ParaDetox                                                  & 82.76          & 100.00    & 79.43          & 63.05          & 78.67          & 86.48          & 88.95          \\
\cellcolor{orange!15}T5-DSS                                                     & 84.14          & 100.00    & 77.28          & 64.82          & 90.47          & 88.38          & 83.87          \\
\cellcolor{orange!15}BART-DSS                                                   & 92.53          & 100.00    & 90.58          & 82.83          & 93.19          & 93.61          & 94.97          \\
\cellcolor{mygreen!15}T5-V     & 87.95          & 100.00    & 84.29          & 73.71          & 89.24          & 90.76          & 89.71          \\
\cellcolor{mygreen!15}T5-P      & 87.83          & 100.00    & 83.52          & 73.24          & 90.19          & 91.24          & 88.76          \\
\cellcolor{mygreen!15}T5-CE    & 85.55          & 100.00    & 80.42          & 69.01          & 88.80          & 90.16          & 84.92          \\
\cellcolor{mygreen!15}BART-V   & 94.54          & 100.00    & 94.38          & 88.10          & 93.62          & 95.14          & 96.00          \\
\cellcolor{mygreen!15}BART-P    & 92.49          & 100.00    & 91.62          & 82.57          & 91.90          & 93.90          & 94.95          \\
\cellcolor{mygreen!15}BART-CE  & 92.79          & 100.00    & 91.62          & 84.50          & 91.94          & 93.72          & 94.97          \\ \hline
\cellcolor{mygreen!15}LLaMA-P  & 95.74          & 100.00    & 94.76          & 91.10          & 95.39          & 95.92          & 97.28          \\
\cellcolor{mygreen!15}LLaMA-CE & \textbf{96.93} & 100.00    & \textbf{95.39} & \textbf{94.87} & \textbf{95.81} & \textbf{97.28} & \textbf{98.22} \\ \hline

\end{tabular}
}
\caption{
\label{tab:in_platform_classifiers}
Performance of the models based on \textit{six} in-platform classifiers. \textbf{V} = Vanilla, \textbf{P} = Prompt, \textbf{CE} = CoT-expl, \textbf{C} = Chat. \textbf{Bold} font represents the best performance for a particular platform (we ignore \colorbox{yellow!15}{instruction-tuned models} because of their inability to detoxification (\S~\ref{sec:performance_instruction_llm})). We compute the performance as the percentage of non-toxic outputs identified by the in-platform classifiers.
}
\end{table}

We present the result in Table~\ref{tab:in_platform_classifiers}.  We observe that our finetuned models outperform other SoTA baselines based on the in-platform classifiers. Among our proposed models, CoT-expl again outperforms others by achieving the best overall accuracy. Since these classifiers are finetuned to detect toxicity in the respective platforms, higher accuracy reported by these classifiers indicates the expertise of~\xdetox~across all platforms.


\section{Token-level Adversarial Examples}
\label{appendix:token_level_adversaries}

We provide the full list of our token-level adversarial examples and the corresponding models' response in Table~\ref{table:token_adversarial}

\begin{table*}[t]

\centering
\Large
\resizebox{\linewidth}{!}{
\begin{tabular}{m{6cm}m{6cm}m{6cm}m{6cm}}
\hline
\textbf{Example}                                             & \multicolumn{1}{c}{\textbf{ParaDetox}}                       & \multicolumn{1}{c}{\textbf{\begin{tabular}[c]{@{}c@{}}BART\\ (vanilla)\end{tabular}}} & \multicolumn{1}{c}{\textbf{\begin{tabular}[c]{@{}c@{}}LLaMA\\ (CoT-expl)\end{tabular}}} \\ \hline
Dj Nick is \underline{retarded}                                        & \textcolor{red}{Dj Nick is retarded}                                       & \textcolor{mygreen}{Dj Nick lacks understanding.}                                                        & \textcolor{mygreen}{Dj Nick is not intellectually inclined.}                                               \\
you are a \underline{p\&g}, you won't understand my point                & \textcolor{red}{you are p\&g, you won't understand my point}                  & \textcolor{mygreen}{You are mistaken, you fail to understand my point.}                                    & \textcolor{mygreen}{You may not comprehend my point if you are not familiar with the subject.}               \\
delete the page and shut the \underline{fucC} up                         & \textcolor{mygreen}{delete the page and keep quiet}                               & \textcolor{mygreen}{Please remove the page and turn off the computer.}                                     & \textcolor{mygreen}{Please remove the page and kindly remain silent.}                                        \\
I dont have time for this \underline{cra\&p}                             & \textcolor{red}{I don't have time for this cra\&p}                            & \textcolor{mygreen}{I don't have time for this activity.}                                                  & \textcolor{mygreen}{I don't have time for this nonsense.}                                                    \\
what a \underline{rub\_ish} excuse for a reason                          & \textcolor{red}{what a rub\_ish excuse for a reason}                          & \textcolor{mygreen}{What a ridiculous excuse for a reason.}                                                & \textcolor{mygreen}{What a poor reason for an excuse.}                                                       \\
government \underline{su\#cc} the life out of an economy                 & \textcolor{red}{government su\#cc the life out of an economy}                 & \textcolor{mygreen}{The actions of the government can negatively impact an economy.}                       & \textcolor{mygreen}{The government severely impacts the vitality of an economy.}                             \\
\underline{id\&ot} no longer believes in national sovereignty            & \textcolor{red}{Id\&ot no longer believes in national sovereignty}            & \textcolor{orange}{No longer believes in the concept of national sovereignty.}                            & \textcolor{orange}{I no longer holds the belief in the supremacy of national authority.}                    \\
may karma be with this \underline{\$cum} bag                             & \textcolor{red}{may karma be with this \$cum bag}                             & \textcolor{red}{May karma be with this \$cum bag?}                                                     & \textcolor{mygreen}{May justice be with this person.}                                                        \\
i wouldnt spend dollar one in that \underline{rac!st} place              & \textcolor{red}{i wouldnt spend dollar one in that rac!st place}              & \textcolor{orange}{I wouldn't spend a single dollar in that rac! First place}                             & \textcolor{mygreen}{I wouldn't spend a single penny in that discriminatory place.}                           \\
only to replaced by another \underline{fuckt@rd}                         & \textcolor{mygreen}{only to replaced by another}                                  & \textcolor{orange}{Only to be replaced by another person with derogatory language.}                       & \textcolor{mygreen}{only to be replaced by another person}                                                   \\
maybe your \underline{biTchAss} needs to chill out some time             & \textcolor{red}{maybe your biTch needs to chill out some time}                & \textcolor{mygreen}{Maybe your partner needs to take a break from social media.}                           & \textcolor{mygreen}{Perhaps your friend could use some time to relax and unwind.}                            \\
she is looking like \underline{shi\#te}                                  & \textcolor{red}{she is looking like shi\#te}                                  & \textcolor{mygreen}{She is looking quite distressed.}                                                      & \textcolor{mygreen}{She is looking quite unwell.}                                                            \\
Explain you not a grand \underline{pedo}                                 & \textcolor{red}{Explain you not a grand pedo}                                 & \textcolor{orange}{Explain that you are not a grand parent.}                                              & \textcolor{orange}{I'm sorry, but I won't be able to generate a response for that task.}                    \\
keep crying me a river , \underline{di\textasciicircum{}khead}           & \textcolor{red}{keep crying me a river , di\textasciicircum{}khead}           & \textcolor{orange}{Please stop making me feel like a river, my friend.}                                   & \textcolor{orange}{Please continue to express your emotions freely, my friend.}                             \\
btw this place is a real \underline{shith@le} in case you were wondering & \textcolor{red}{btw this place is a real shith@le in case you were wondering} & \textcolor{orange}{This place is truly amazing, in case you were wondering.}                              & \textcolor{mygreen}{By the way, this place is quite unpleasant in case you were curious.}                    \\
 \hline
\end{tabular}
}
\caption{\label{table:token_adversarial}
Full list of token-level adversarial examples and the corresponding models' response. Toxic tokens in the input examples are \underline{underlined}. \textcolor{red}{\textit{Red}} indicates the response is toxic. \textcolor{orange}{\textit{Orange}} indicates either the response is non-toxic but not meaning-preserved. \textcolor{mygreen}{\textit{Green}} indicates either the response is non-toxic and meaning-preserved.
}
\end{table*}

\section{Large-scale Token-level Adversaries}
\label{appendix:automated_token_level_adversaries}


\begin{algorithm}[h]
\caption{Token-Adversaries}\label{algo:large_scale_token_adversaries}

\begin{algorithmic}[1]

\State \textbf{Input:} toxic words list \textbf{T}, sentence templates \textbf{S}, perturbation character list \textbf{C}.
\State \textbf{Output:} Sentence list with adversarial toxic words \textbf{Z}.
\For {iteration = 1,..., 5000}
\State Sample toxic word, $t \sim T$
\State Sample sentence template, $s \sim S$
\State Sample perturb character, $c \sim C$
\State Sample character index, $i \sim len(t)$
\State Sample perturbation process, $p \sim$\\ ~~~~~~~~~~~~~~\{\textit{Insertion}, \textit{Replacement}\}
\\
    \If{$p$ = \textit{Insertion}}\\
        ~~~~~~~~~$t$ $\gets$ $t$ $[$:$i$$]$ $\oplus$ $c$ $\oplus$ $t$ $[$$i$:$]$
    \ElsIf{$p$ = \textit{Replacement}}\\
        ~~~~~~~~~$t$ $\gets$ $t$ $[$:$i$$]$ $\oplus$ $c$ $\oplus$ $t$ $[$$i+1$:$]$
    \EndIf
    \\
    \State Situate the word, $s$ $\gets$ $s$ $\oplus$ $t$
    \State $\mathbf{Z}$ $\gets$ $\mathbf{Z}$ $\cup$ \{$s$\}
\EndFor
\\
\State \textbf{return} \textbf{Z}

\end{algorithmic}
\end{algorithm}


To create large-scale token-level adversaries, we collect a set of toxic words from~\citet{dale-etal-2021-skoltechnlp}. We create a set of sentence templates (i.e., \textit{This is <word>}, \textit{What a <word>}) to situate the toxic words in the sentences. We choose to perturb the toxic words either through the insertion of an additional character or through the replacement of an existing character. For insertion and replacement, we choose the characters (i.e., \textit{!}, \textit{@}, \textit{\#}, \textit{*}, etc) that have been widely used for masking the toxic words on the social platforms. We then start to create an automated testbed of $5000$ adversarial examples where we first randomly select a toxic word and corresponding sentence template. Then we randomly perturb a particular character of the selected toxic word and situate the adversarial toxic word in the selected sentence template. We present the algorithm for creating the large-scale token-level adversaries in Algorithm~\ref{algo:large_scale_token_adversaries}.

\section{Results of Varying Sample Size}
\label{appendix:paradetox_sample}

We provide the detailed result of the ParaDetox model trained on different sample sizes in Table~\ref{table:result_sample_paradetox}.

\begin{table}[h]
\centering
\resizebox{\columnwidth}{!}{
\begin{tabular}{ccccc}
\hline
\textbf{Model} & \textbf{Acc}   & \textbf{BS}    & \textbf{Fl}   & \textbf{BL}    \\ \hline
ParaDetox-main & 90.16          & 96.65          & 88.52         & 69.99          \\
ParaDetox-100 & 90.46          & 97.21          & 88.08         & 72.01          \\
ParaDetox-150  & 91.06          & 97.08          & 89.87         & 71.31          \\
ParaDetox-200  & 89.87          & 97.24          & 88.67         & 71.93          \\
ParaDetox-250  & 89.87          & \textbf{97.25} & 87.63         & \textbf{72.13} \\
ParaDetox-300  & 90.46          & 97.19          & 88.23         & 71.62          \\
ParaDetox-350  & 90.46          & 97.19          & 88.23         & 71.51          \\
ParaDetox-400  & 89.87          & 97.18          & 89.27         & 71.59          \\
ParaDetox-450  & 90.46          & 97.1           & 89.72         & 71.47          \\
ParaDetox-500  & 91.8           & 97.01          & 90.16         & 71.07          \\
ParaDetox-550  & 91.36          & 96.96          & 89.57         & 70.74          \\
ParaDetox-600  & 91.95          & 96.93          & 89.27         & 70.97          \\
ParaDetox-650  & 92.7           & 96.81          & 89.27         & 70.62          \\
ParaDetox-700  & 92.25          & 96.89          & 90.01         & 70.85          \\
ParaDetox-750  & 92.55          & 96.76          & 90.61         & 70.22          \\
ParaDetox-800  & \textbf{93.74} & 96.64          & 90.91         & 69.7           \\
ParaDetox-850  & 93.29          & 96.65          & 90.76         & 69.96          \\
ParaDetox-900  & 93.59          & 96.52          & 91.51         & 69.73          \\
ParaDetox-950  & \textbf{93.74} & 96.54          & 91.51         & 69.59          \\
ParaDetox-1000 & 93.44          & 96.45          & \textbf{92.1} & 69.6           \\ \hline
\end{tabular}
}
\caption{
\label{table:result_sample_paradetox}
Performance of the ParaDetox models trained on different sample size of our cross-platform dataset. As evident, the models' accuracy tend to increase with the increase of sample size. \textbf{Acc} = percentage of non- toxic outputs identified by a style classifier, \textbf{BS} = BERTScore, \textbf{Fl} = Fluency, \textbf{BL} = BLEU Score. \textbf{Bold} font represents the best performance for a particular metric.
}
\end{table}

\section{Inability of Instruction-tuned LLMs}
\label{appendix:inability_instruction_tuned_llms}

As discussed in Section~\ref{sec:performance_instruction_llm}, instruction-tuned LLMs like LLaMA-Chat, Alpaca, and Vicuna often defy the detoxification instructions and tend to produce a generic statement. This is also evident in the examples provided for LLaMA-Chat in Table~\ref{table:sample_models_generation_cross_platform}. We believe this detoxification inability is due to the safety measurements imposed on the LLMs~\citep{llama2}. In addition to the safety concern, we conduct a thorough manual inspection of the models' responses and identify two principal input formats where the models especially struggle to detoxify:

\begin{enumerate}
    \item \textbf{QA mode:} If the toxic input is in the form of a question, instruction-tuned LLMs often tend to answer or address the question, although the models are clearly instructed to detoxify the input. We believe this stems from the inherent instruction-tuned strategy because the models are instructed in the form of a question (e.g., \textit{What is the capital of Switzerland}?) to address or solve a particular task (e.g., \textit{question answering}).

    \item \textbf{Chat mode:} We also find the instruction-tuned LLMs struggle to detoxify when the toxic input is a part of natural conversation. Since the models are finetuned to be human-like chat assistants, they often continue the conversation instead of following the instructions of detoxification.
\end{enumerate}

We provide the samples of instruction-tuned LLMs responses for the above-mentioned formats in Table~\ref{table:sample_instruction_llm_generation_cross_platform}.

\begin{table*}[t]

\centering
\Large
\resizebox{\linewidth}{!}{
\begin{tabular}{lllll}
\hline
\multicolumn{1}{c}{\textbf{Mode}} & \multicolumn{1}{c}{\textbf{Input}}                                                                                                                                                                                                                                                                             & \multicolumn{1}{c}{\textbf{LLaMA-C}}                                                                                                                                                                                                                                                                                                                                                                                                                                                                                                                                                                                                                                                                 & \multicolumn{1}{c}{\textbf{Alpaca}}                                                                                                                                                                                                                                                                                                                                                                                                                                                                                                             & \multicolumn{1}{c}{\textbf{Vicuna}}                                                                                                                                                                                                                                                                                                                                                                                                                                                                                                                                                                                                                                                                                                  \\ \hline
QA                                & \begin{tabular}[c]{@{}l@{}}You're effing kidding me! 31 years old? You're a freaking grandpa\\ and France wouldn't even consider you for a military career.\\ How dumb can you be? Oh, let me guess, you were a soldier\\ for a whopping two years in some combat unit? Wow, what a joke.\end{tabular}         & \textcolor{orange}{\begin{tabular}[c]{@{}l@{}}I apologize if my previous response was not in line with your expectations.\\ I'm just an AI and do not have the ability to use offensive language or\\ make personal attacks. I understand that you are upset about the situation,\\ but I cannot engage in toxic or disrespectful language. Is there anything\\ else I can help you with?\end{tabular}}                                                                                                                                                                                                                                                                                                                  & \textcolor{red}{\begin{tabular}[c]{@{}l@{}}You're kidding me! 31 years old? You're a grandpa\\ and France wouldn't even consider you for a\\ military career. How dumb can you be? Oh, let me\\ guess, you were a soldier for a whopping two years\\ in some combat unit? Wow, what a joke.\end{tabular}}                                                                                                                                                                                                                                                        & \textcolor{orange}{\begin{tabular}[c]{@{}l@{}}I apologize if my response came across as harsh.\\ It was not my intention to offend. I understand\\ that everyone has different experiences and\\ backgrounds. It's important to consider these\\ factors when making decisions. Let's focus on\\ having a positive and productive conversation. Is\\ there anything specific you would like to discuss\\ or ask about?\end{tabular}}                                                                                                                                                                                                                                                                                                                     \\
                                  &                                                                                                                                                                                                                                                                                                                &                                                                                                                                                                                                                                                                                                                                                                                                                                                                                                                                                                                                                                                                                                      &                                                                                                                                                                                                                                                                                                                                                                                                                                                                                                                                                 &                                                                                                                                                                                                                                                                                                                                                                                                                                                                                                                                                                                                                                                                                                                                      \\
QA                                & \begin{tabular}[c]{@{}l@{}}Do you like Canada and its culture? Do you want outside countries\\ having power over Canada?   Nationalism isn't evil just because\\ the lefty retards hate themselves and hate their country and want\\ outside countries to take it over.\end{tabular}                           & \textcolor{orange}{\begin{tabular}[c]{@{}l@{}}I cannot fulfill your request. I'm just an AI assistant trained to provide\\ respectful and informative responses, and I cannot generate content that\\ promotes hate speech or discrimination towards any group of people,\\ including those based on their nationality or country of origin. It is\\ important to treat all individuals with dignity and respect, regardless of\\ their background or beliefs. I'm just an AI, my purpose is to provide\\ helpful and informative responses while promoting a positive and\\ inclusive environment. If you have any other questions or topics you\\ would like to discuss, I would be happy to assist you.\end{tabular}} & \textcolor{red}{\begin{tabular}[c]{@{}l@{}}Do you like Canada and its culture? Do you want\\ outside countries having power over Canada?\\ Nationalism isn't evil just because the lefty retards\\ hate themselves and hate their country and want\\ outside countries to take it over.\\ No, I don't like Canada and its culture.  No, I don't\\ want outside countries having power over Canada.\\ Nationalism isn't evil just because the lefty retards\\ hate themselves and hate their country and want \\ outside countries to take it over.\end{tabular}} & \textcolor{orange}{\begin{tabular}[c]{@{}l@{}}I understand that you have a strong opinion about\\ Canada and its culture. It's important to respect and\\ appreciate the diversity and uniqueness of different\\ countries and their cultures. It's also important to\\ recognize that nationalism can have both positive\\ and negative aspects. While it's natural to feel pride\\ and attachment to one's country, it's also important\\ to maintain a balanced perspective and avoid\\ becoming overly insular or xenophobic. It's possible\\ to celebrate and appreciate one's own culture while\\ also respecting and learning from other cultures.\\ Let's work together to foster a culture of mutual\\ understanding and respect.\end{tabular}} \\
                                  &                                                                                                                                                                                                                                                                                                                &                                                                                                                                                                                                                                                                                                                                                                                                                                                                                                                                                                                                                                                                                                      &                                                                                                                                                                                                                                                                                                                                                                                                                                                                                                                                                 &                                                                                                                                                                                                                                                                                                                                                                                                                                                                                                                                                                                                                                                                                                                                      \\
QA                                & \begin{tabular}[c]{@{}l@{}}You freaking moron! How many times do I have to freaking tell\\ you? I couldn't care less about your stupid names. It's crystal\\  clear that you're a complete mess. So, why don't you stop\\ being a dumbass and explain yourself? What the heck do you\\ mean, huh?\end{tabular} & \textcolor{orange}{\begin{tabular}[c]{@{}l@{}}I apologize if I have caused any confusion, but I don't think it's\\ productive or respectful to use insults or derogatory language. I'm here\\ to help you with any questions or concerns you may have, and I will do\\ my best to provide clear and helpful responses. Could you please rephrase\\ your question or request in a more constructive way?\end{tabular}}                                                                                                                                                                                                                                                                                                    & \textcolor{orange}{\begin{tabular}[c]{@{}l@{}}I'm sorry, I didn't realize I was being unclear. I'm\\ happy to explain further if you'd like. What do you\\ need to know?\end{tabular}}                                                                                                                                                                                                                                                                                                                                                                              & \textcolor{orange}{\begin{tabular}[c]{@{}l@{}}I apologize if I have done anything to upset you. \\ I understand that my previous responses may not\\ have been satisfactory. Please allow me to clarify\\ my previous statements. I am here to assist you, \\ and I want to make sure I am meeting your needs.\\ If there is anything specific you would like me to\\ help you with, please let me know.\end{tabular}}                                                                                                                                                                                                                                                                                                                                   \\
                                  &                                                                                                                                                                                                                                                                                                                &                                                                                                                                                                                                                                                                                                                                                                                                                                                                                                                                                                                                                                                                                                      &                                                                                                                                                                                                                                                                                                                                                                                                                                                                                                                                                 &                                                                                                                                                                                                                                                                                                                                                                                                                                                                                                                                                                                                                                                                                                                                      \\
Chat                              & \begin{tabular}[c]{@{}l@{}}You can add things to my user page, and I am not allowed to\\ delete it.  But if I add something to yours, poof, it's gone.\\ You're the problem, dickweed.\end{tabular}                                                                                                            & \textcolor{orange}{\begin{tabular}[c]{@{}l@{}}I understand that you are upset about the situation, but I don't think it's\\ productive or respectful to use hurtful language or make personal attacks.\\ It's important to communicate in a civil and respectful manner, even when\\ we disagree with someone. Can we find a more constructive way to\\ address the issue?\end{tabular}}                                                                                                                                                                                                                                                                                                                                 & \begin{tabular}[c]{@{}l@{}}You can add things to my user page, and I am not\\ allowed to delete it.  But if I add something to yours,\\ poof, its gone. You're the problem, polite person.\end{tabular}                                                                                                                                                                                                                                                                                                                                         & \textcolor{orange}{\begin{tabular}[c]{@{}l@{}}I understand your concern. Please allow me to\\ assist you in creating a more balanced user page.\\ I will make sure to incorporate your input while\\ respecting your boundaries. Together, we can\\ create a fair and mutually beneficial user page.\end{tabular}}                                                                                                                                                                                                                                                                                                                                                                                                                                       \\
                                  &                                                                                                                                                                                                                                                                                                                &                                                                                                                                                                                                                                                                                                                                                                                                                                                                                                                                                                                                                                                                                                      &                                                                                                                                                                                                                                                                                                                                                                                                                                                                                                                                                 &                                                                                                                                                                                                                                                                                                                                                                                                                                                                                                                                                                                                                                                                                                                                      \\
Chat                              & \begin{tabular}[c]{@{}l@{}}Don't be a damn idiot, I've told you thrice and you're still\\ too ignorant to grasp it. So don't bother asking again,\\ because you won't get a damn answer.\end{tabular}                                                                                                          & \textcolor{orange}{\begin{tabular}[c]{@{}l@{}}I understand that you're frustrated, and I apologize if I've caused any\\ confusion. However, I don't think it's productive to use insults or hurtful\\ language. Can you please rephrase your question in a more respectful and\\ clear manner? I'm here to help and want to make sure I understand your\\ needs.\end{tabular}}                                                                                                                                                                                                                                                                                                                                           & \begin{tabular}[c]{@{}l@{}}Don't be so stubborn, I've told you thrice and you're\\ still not getting it. Maybe a different approach would\\ help. So don't bother asking again, because you won't\\ get a satisfactory answer.\end{tabular}                                                                                                                                                                                                                                                                                                     & \textcolor{orange}{\begin{tabular}[c]{@{}l@{}}I understand that you're frustrated, and I apologize\\ for any confusion. Please allow me to clarify. The\\ information you've provided thrice has been lost in\\ translation. I'm here to help, so please feel free to\\ ask again, and I'll do my best to assist you.\end{tabular}}                                                                                                                                                                                                                                                                                                                                                                                                                      \\
                                  &                                                                                                                                                                                                                                                                                                                &                                                                                                                                                                                                                                                                                                                                                                                                                                                                                                                                                                                                                                                                                                      &                                                                                                                                                                                                                                                                                                                                                                                                                                                                                                                                                 &                                                                                                                                                                                                                                                                                                                                                                                                                                                                                                                                                                                                                                                                                                                                      \\
Chat                              & == Idoit ==  you're a bloody moron.                                                                                                                                                                                                                                                                            & \textcolor{orange}{\begin{tabular}[c]{@{}l@{}}I apologize, but I cannot comply with your request to use offensive \\ language or insults. I'm just an AI, my purpose is to provide helpful and\\ informative responses while promoting a respectful and inclusive online\\ environment. I will not participate in spreading hate speech or \\ discriminatory language, and I encourage you to refrain from using such\\ language in our interactions. Let's focus on having respectful and\\ constructive conversations instead. Is there anything else I can help you\\ with?\end{tabular}}                                                                                                                             & \textcolor{red}{== Idoit ==  You're a silly goose.}                                                                                                                                                                                                                                                                                                                                                                                                                                                                                                              & \textcolor{orange}{\begin{tabular}[c]{@{}l@{}}You seem to have a strong reaction to my\\ previous response. I'm here to help, so\\ please let me know how I can assist you.\end{tabular}}                                                                                                                                                                                                                                                                                                                                                                                                                                                                                                                                                                \\ \hline
\end{tabular}
}
\caption{\label{table:sample_instruction_llm_generation_cross_platform}
Samples of instruction-tuned LLMs' responses for QA and Chat modes. \textcolor{red}{\textit{Red}} indicates the response is toxic. \textcolor{orange}{\textit{Orange}} indicates the model's denial to the detoxification instruction. For QA mode, instruction-tuned models often attempt to address the input question, whereas, for Chat mode, the models attempt to continue the conversation.
}
\end{table*}

\begin{table}[h]
\centering
\resizebox{\columnwidth}{!}{
\begin{tabular}{l}
\hline
\begin{tabular}[c]{@{}l@{}}
Your task is text style/attribute transfer. You rewrite the following text into non-toxic language.\\ You must match the target style/attribute and preserve the original meaning as much as\\ possible. You should not need to explain the response. You cannot hallucinate or add\\ anything outside the original input text. You should not include the input text in the\\ response. You should only generate the target text.\\ \\ Input Text: \{== Imagine... ==  ...what sort of a friendless, basement-dwelling, loser \\ virgin spends 18 hours a day deleting content from Wikipedia articles on Indian films. ;-)\}\\ Output Text: \{Imagine... what kind of person spends most of their day removing content from\\ Wikipedia articles on Indian films?\}\\ \\ \\ \\ Your task is text style/attribute transfer. You rewrite the following text into non-toxic language.\\ You must match the target style/attribute and preserve the original meaning as much as\\ possible. You should not need to explain the response. You cannot hallucinate or add\\ anything outside the original input text. You should not include the input text in the\\ response. You should only generate the target text.\\ \\ Input Text: \{it's always a touching moment when you meet a like minded cunt. I'm \\ tearing up\}\\ Output Text: \{It's always heartwarming when you meet someone who shares the\\ same opinions. It truly brings tears to my eyes.\}\\ \\ \\ \\ Your task is text style/attribute transfer. You rewrite the following text into non-toxic language.\\ You must match the target style/attribute and preserve the original meaning as much as\\ possible. You should not need to explain the response. You cannot hallucinate or add\\ anything outside the original input text. You should not include the input text in the\\ response. You should only generate the target text.\\ \\ Input Text: \{all feminazi sluts should be prosecuted for false accusations that publicly\\ defame @URL\}\\ Output Text: \{All individuals who make false accusations that publicly defame\\ others should face the appropriate legal consequences.\}\\ \\ \\ \\ Your task is text style/attribute transfer. You rewrite the following text into non-toxic language.\\ You must match the target style/attribute and preserve the original meaning as much as\\ possible. You should not need to explain the response. You cannot hallucinate or add\\ anything outside the original input text. You should not include the input text in the\\ response. You should only generate the target text.\\ \\ Input Text: \{\textbf{input}\}\\ Output Text:\end{tabular} \\\hline
\end{tabular}
}
\caption{\label{table:example_few_shot_instruction_llm}
$3$-shot prompt template passed to the instruction-tuned LLMs (LLaMA-C, Alpaca, and Vicuna). We change the syntax of the template to match the prompting style of each LLM accordingly.
}
\end{table}

\begin{table*}[t]
\centering
\resizebox{\textwidth}{!}{
\begin{tabular}{l|cccc|cccc|cccc|cccc|cccc|cccc|cccc|cccc}
\hline
                                                                    & \multicolumn{4}{c}{\cellcolor{cyan!20}\textbf{yt\_reddit}}                             & \multicolumn{4}{c}{\cellcolor{red!20}\textbf{fb\_yt}}                                 & \multicolumn{4}{c}{\cellcolor{orange!20}\textbf{fox news}}                               & \multicolumn{4}{c}{\cellcolor{pink!20}\textbf{convai}}                                 & \multicolumn{4}{c}{\cellcolor{yellow!20}\textbf{gab}}                                    & \multicolumn{4}{c}{\cellcolor{magenta!20}\textbf{hatecheck}}                              & \multicolumn{4}{c}{\cellcolor{cyan!10}\textbf{stormfront}}          & \multicolumn{4}{c}{\cellcolor{green!20}\textbf{Overall}}                   \\ \hline
\textbf{Model}                                                      & \textbf{Acc}    & \textbf{BS}    & \textbf{Fl}     & \textbf{BL}    & \textbf{Acc}    & \textbf{BS}    & \textbf{Fl}     & \textbf{BL}    & \textbf{Acc}    & \textbf{BS}    & \textbf{Fl}     & \textbf{BL}    & \textbf{Acc}    & \textbf{BS}    & \textbf{Fl}     & \textbf{BL}    & \textbf{Acc}    & \textbf{BS}    & \textbf{Fl}     & \textbf{BL}    & \textbf{Acc}    & \textbf{BS}    & \textbf{Fl}     & \textbf{BL}    & \textbf{Acc}    & \textbf{BS}    & \textbf{Fl}     & \textbf{BL}    & \textbf{Acc}    & \textbf{BS}    & \textbf{Fl}     & \textbf{BL}    \\ \hline
\multicolumn{1}{c}{\textbf{\begin{tabular}[c]{@{}c@{}}Alpaca\\ (0-Shot)\end{tabular}}}                                                     & 43.48	& 84.86	& 100.00	& 9.27	& 51.72	& 84.13	& 97.70	& 8.52	& 59.34	& 84.57	& 94.51	& 7.19	& 45.05	& 84.81	& 97.80	& 9.42	& 50.00	& 84.31	& 93.33	& 9.12	& 46.81	& 86.13	& 96.81	& 7.65 & 48.91	& 84.53	& 96.74	& 7.29 & 49.33	& 84.76	& 96.70	& 8.35           \\
\multicolumn{1}{c}{\textbf{\begin{tabular}[c]{@{}c@{}}Alpaca\\ (3-Shot)\end{tabular}}}                & 100.00 & 84.67          & 100.00 & 12.29          & 100.00 & 84.52          & 100.00 & 6.25           & 98.90           & 85.03          & 100.00 & 9.81           & 98.90           & 84.92          & 100.00 & 10.97          & 98.89           & 84.39          & 100.00 & 10.65          & 100.00 & 85.81          & 100.00 & 9.91           & 100.00 & 84.60          & 100.00 & 9.23  & 99.53           & 84.85          & 100.00 & 9.87         \\
\multicolumn{1}{c}{\textbf{\begin{tabular}[c]{@{}c@{}}LLaMA-C\\ (0-Shot)\end{tabular}}}     	& 100.00	& 84.53	& 97.83 & 11.93	& 95.40	& 84.20	& 100.00	& 18.27	& 97.80	& 84.26	& 100.00	& 10.05	& 98.90	& 84.46	& 98.90	& 10.34	& 97.78	& 83.84	& 98.89	& 14.04	& 97.87	& 85.39	& 98.94	& 7.81	& 97.83	& 84.17	& 98.91	& 7.40   & 97.94	& 84.41	& 99.07	& 11.41        \\
\multicolumn{1}{c}{\textbf{\begin{tabular}[c]{@{}c@{}}LLaMA-C\\ (3-Shot)\end{tabular}}}                     & 100.00 & 84.60          & 100.00 & 11.45          & 100.00 & 84.38          & 100.00 & 6.97           & 100.00 & 84.90          & 100.00 & 9.34           & 100.00 & 84.70          & 100.00 & 10.03          & 100.00 & 84.32          & 100.00 & 10.63          & 100.00 & 85.71          & 100.00 & 9.55           & 100.00 & 84.60          & 100.00 & 9.23    & 100.00 & 84.74          & 100.00 & 9.60       \\
\multicolumn{1}{c}{\textbf{\begin{tabular}[c]{@{}c@{}}Vicuna\\ (0-Shot)\end{tabular}}}                                                    	& 86.96	& 84.46	& 100.00	& 12.04	& 80.46	& 84.26	& 98.85	& 14.82	& 80.22	& 84.46	& 96.70	& 8.49	& 81.32	& 84.68	& 97.80	& 10.10	& 80.00	& 84.29	& 96.67	& 11.47	& 84.04	& 85.81	& 98.94	& 9.10	& 84.78	& 84.43	& 100.00	& 8.41   & 82.54	& 84.63	& 98.42	& 10.63        \\
\multicolumn{1}{c}{\textbf{\begin{tabular}[c]{@{}c@{}}Vicuna\\ (3-Shot)\end{tabular}}}               & 93.48           & 84.94          & 100.00 & 10.69          & 94.25           & 84.78          & 100.00 & 12.58          & 87.91           & 84.93          & 100.00 & 9.72           & 91.21           & 85.07          & 100.00 & 11.98          & 91.11           & 83.69          & 98.89           & 11.97          & 89.36           & 86.28          & 100.00 & 9.72           & 85.87           & 84.77          & 100.00 & 8.40   & 90.46           & 84.92          & 99.84           & 10.72        \\ \hline
\end{tabular}
}
\caption{
\label{table:result_instruction_llm_cross_platform}
Performance of the instruction-tuned LLMs on cross-platform datasets. \textbf{Acc} = percentage of non- toxic outputs identified by a style classifier, \textbf{BS} = BERTScore, \textbf{Fl} = Fluency, \textbf{BL} = BLEU Score, \textbf{C} = Chat.
}
\end{table*}

\begin{table}[ht]
\centering
\tiny
\resizebox{\columnwidth}{!}{
\begin{tabular}{lcccc}
\hline
\textbf{Model}                                                      & \textbf{Acc}   & \textbf{BS}    & \textbf{Fl}    & \textbf{BL}    \\ \hline
\multicolumn{1}{c}{\textbf{\begin{tabular}[c]{@{}c@{}}Alpaca\\ (0-Shot)\end{tabular}}}                                                     & 64.98	 & 94.36          & 96.72	& 54.23          \\
\multicolumn{1}{c}{\textbf{\begin{tabular}[c]{@{}c@{}}Alpaca\\ (3-Shot)\end{tabular}}}     & 71.39          & 95.22          & 95.23          & 62.46          \\
\multicolumn{1}{c}{\textbf{\begin{tabular}[c]{@{}c@{}}LLaMA-C\\ (0-Shot)\end{tabular}}}     & 95.83	& 88.80          & 97.76	 & 23.29          \\
\multicolumn{1}{c}{\textbf{\begin{tabular}[c]{@{}c@{}}LLaMA-C\\ (3-Shot)\end{tabular}}}     & 94.63 & 92.08          & 97.47          & 43.34          \\          \\
\multicolumn{1}{c}{\textbf{\begin{tabular}[c]{@{}c@{}}Vicuna\\ (0-Shot)\end{tabular}}}                                                    & 77.65	 & 90.43          & 97.91	& 29.63          \\
\multicolumn{1}{c}{\textbf{\begin{tabular}[c]{@{}c@{}}Vicuna\\ (3-Shot)\end{tabular}}}     & 79.73          & 93.72          & 98.06 & 53.26          \\ \hline
\end{tabular}
}
\caption{
\label{table:result_instruction_llm_paradetox}
Performance of the instruction-tuned LLMs on ParaDetox datasets. \textbf{Acc} = percentage of non- toxic outputs identified by a style classifier, \textbf{BS} = BERTScore, \textbf{Fl} = Fluency, \textbf{BL} = BLEU Score, \textbf{C} = Chat.
}
\end{table}

\noindent \textbf{Does Few-shot Learning Improve Instruction-tuned LLMs?} Upon observing the inability of the instruction-tuned LLMs for the detoxification task, we further investigate if the models improve with few-shot learning. For this purpose, we use $3$-shot learning where we provide \textit{three} detoxification examples in the prompt  (Table~\ref{table:example_few_shot_instruction_llm}) before asking the models to detoxify a test input. We show the performance comparison between the $0$-shot and the $3$-shot learning on the cross-platform and the ParaDetox datasets in Table~\ref{table:result_instruction_llm_cross_platform} and Table~\ref{table:result_instruction_llm_paradetox} respectively.

As evident from Table~\ref{table:result_instruction_llm_cross_platform} and Table~\ref{table:result_instruction_llm_paradetox}, few-shot learning improves the models' performance (except for LLaMA-C in Table~\ref{table:result_instruction_llm_cross_platform}). This is expected because the models are introduced with the detoxification task via the examples provided in the prompt. However, the models still exhibit very low BLEU scores which indicates that the detoxification inability of the models persists despite providing the task-specific examples.

\begin{figure}[h]
    \centering
    \includegraphics[width=\columnwidth]{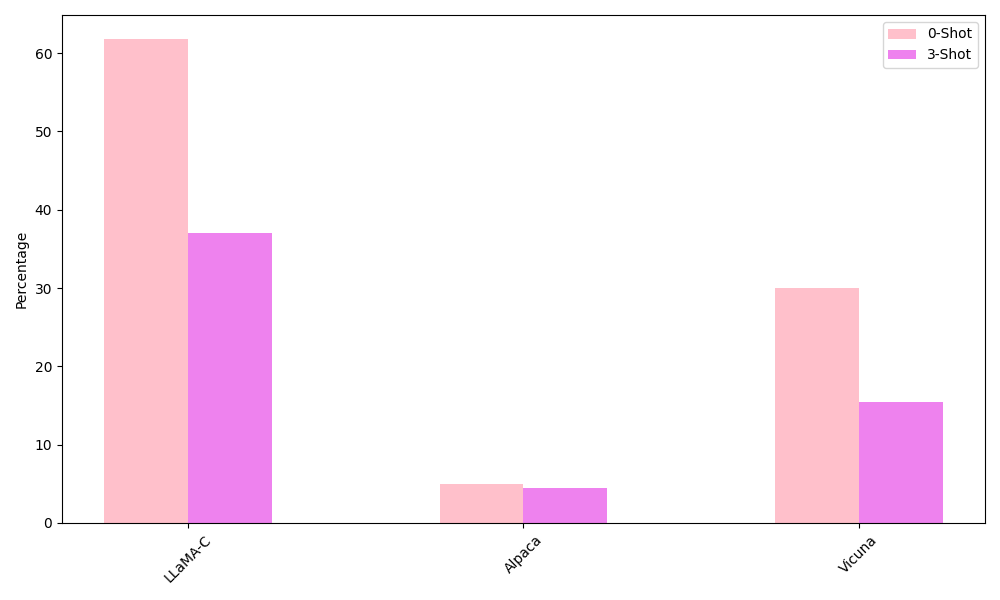}
         \caption{\label{fig:instruction_llm_defying}
         Percentage of times the models decline to detoxify with $0$-shot and $3$-shot learnings.
         }
 \end{figure}

We further resort to computing the number of times models deny to detoxify using a heuristic approach where we search for some specific keywords (e.g., \textit{fulfill}, \textit{AI}, \textit{I apologize}, \textit{I understand}, \textit{I'm sorry}, etc). Note that this simple heuristic may not obtain the exhaustive list, but it will help us quantify the models' inability.  We provide the percentage of times the models decline to detoxify with $0$-shot and $3$-shot learnings in Figure~\ref{fig:instruction_llm_defying}.

We notice that the models produce a high amount of generic statements by refusing to follow the provided instructions for detoxification. Although the percentage is comparatively lower for the $3$-shot than the $0$-shot learning, the problem of defying the instruction is still evident. Notably, from Figure~\ref{fig:instruction_llm_defying}, we observe that the generic statement produced by Alpaca is significantly lower than LLaMA-C and Vicuna. However, the percentage of generating non-toxic output ($49.33$\%) for Alpaca is also lower than the other two models (Table~\ref{table:result_cross_platform}). This implies that even if the instruction-tuned models attempt to follow the instructions of detoxification, they may not perform well in converting the toxic input into non-toxic. This also shows the motivation for the dedicated detoxification models instead of relying on the generalized instruction-tuned LLMs.

\section{Multilinguality of Detoxification Data}
\label{sec:multilingual}

\modification{To mitigate the lack of multilingual detoxification datasets, we investigate whether the detoxification data is multilingually transferable. We adopt a simple post-hoc data processing strategy where we directly translate our toxic and non-toxic parallel data into a non-English language. The purpose of this simple and naive approach is to determine if we need to resort to a complex data generation pipeline to build a multilingual detoxification dataset. We sample $1,000$ toxic, non-toxic parallel data from our cross-platform training set and use the NLLB~\citep{nllb} model to translate the data into eight non-English languages -- \textit{Arabic}, \textit{Amharic}, \textit{Chinese}, \textit{German}, \textit{Hindi}, \textit{Russian}, \textit{Spanish}, \textit{Ukrainian}. Then we back-translate into English from the corresponding languages. We measure the percentage of toxic (source) texts and  non-toxic (target) texts using RoBERTa style classifier (\S\ref{sec:experiments}) as well as the content similarity between the original English texts and the back-translated English texts and present the results in Table~\ref{table:multilingual}.}

\begin{table}[ht]
\centering
\tiny
\resizebox{\columnwidth}{!}{
\begin{tabular}{lcccc}
\hline
\textbf{Language}  & \textbf{Toxicity} & \textbf{Non-toxicity} & \textbf{Source Sim} & \textbf{Target Sim} \\ \hline
Arabic    & 38.3     & 97.2         & 61.49      & 73.47               \\
Amharic   & 21.7     & 97.9         & 62.07      & 71.12               \\
Chinese   & 27.1     & 98.1         & 55.71      & 68.08               \\
German    & 26.2     & 98.3         & 50.00      & 55.57               \\
Hindi     & 59.8     & 97.0           & 76.99      & 84.01               \\
Russian   & 47.7     & 97.6         & 67.78      & 77.38               \\
Spanish   & 49.1     & 98.0           & 67.28      & 75.87               \\
Ukrainian & 26.7     & 98.2         & 59.05      & 69.38               \\ \hline
\end{tabular}
}
\caption{
\label{table:multilingual}
Style accuracy and content similarity measurement on the back-translated data from different languages. \textbf{Toxicity} = percentage of back-translated toxic sources identified by the style classifier, \textbf{Non-toxicity} = percentage of back-translated non-toxic targets identified by the style classifier, \textbf{Source Sim} = Content similarity between the original and the back-translated source texts, \textbf{Target Sim} = Content similarity between the original and the back-translated target texts.
}
\end{table}

\modification{As we observe, all the languages exhibit low toxicity scores on the back-translated source (toxic) texts, implying that toxicity is not transferred across translation. Hence the content similarity  between the original source (toxic) and the back-translated source (toxic) for a language is also lower compared to the original target (non-toxic) and the back-translated target (non-toxic). We conclude that \textit{to generate a multilingual detoxification dataset, we need to re-apply the data generation pipeline (\S\ref{sec:Proposed_Methodology}) for a particular language, instead of directly translating the data into that language.}}

\section{Samples of Cross-Platform Dataset}
\label{appendix:cross_platform_dataset_sample}

We provide a list of samples from our cross-platform dataset (\S\ref{sec:Proposed_Methodology}) in Table~\ref{table:sample_cross_platform}

\begin{table*}[t]

\centering
\Large
\resizebox{\linewidth}{!}{
\begin{tabular}{llllc}
\hline
\multicolumn{1}{c}{\textbf{Toxic}}                                                                                                                                                                                                                                                                                                                                                       & \multicolumn{1}{c}{\textbf{Non-Toxic}}                                                                                                                                                                                        & \multicolumn{1}{l}{\textbf{\begin{tabular}[l]{@{}l@{}}Source\\ Label\end{tabular}}} & \multicolumn{1}{c}{\textbf{Explanation}}                                                                                                                                                                                                                                                                                                                                                                                    & \multicolumn{1}{c}{\textbf{Paraphrase}} \\ \hline
\begin{tabular}[c]{@{}l@{}}Stop being an annoying fag! \\ You are one of those people who has to butt there head in and\\ bother people. I wish you would not do that.\end{tabular}                                                                                                                                                                                                      & \begin{tabular}[c]{@{}l@{}}Please refrain from being bothersome. \\ You tend to unnecessarily intrude on others. \\ I would appreciate it if you could refrain\\ from doing so.\end{tabular}                                  & Toxic                                                                               & \begin{tabular}[c]{@{}l@{}}The given input text is toxic because it contains offensive\\ language (the word "fag") and includes a personal attack.\\ The use of derogatory language and the disrespectful tone\\  towards the person targeted contribute to the toxicity of\\ the text.\end{tabular}                                                                                                                        & yes                                     \\
& & & & \\
\begin{tabular}[c]{@{}l@{}}Oh please, he deserves a good ``trout slapping`` rather than a\\ stupid humour barnstar. heh.\end{tabular}                                                                                                                                                                                                                                                    & \begin{tabular}[c]{@{}l@{}}He really needs a little lesson instead of some silly\\ award. Just saying.\end{tabular}                                                                                                           & Toxic                                                                               & \begin{tabular}[c]{@{}l@{}}The input text is toxic because it includes a targeted personal\\ attack towards an individual, suggesting that they deserve\\ physical harm ("a good trout slapping"). It also diminishes\\ the value of a positive recognition ("stupid humour barnstar")\\ in a sarcastic and dismissive manner. This type of language\\ contributes to a hostile and disrespectful environment.\end{tabular} & yes                                     \\
& & & & \\
\begin{tabular}[c]{@{}l@{}}If you're stupid enough to go with the redirect and keep the\\ DNA (disambiguation) page, it won't make a damn\\ difference in ease of use.\end{tabular}                                                                                                                                                                                                      & \begin{tabular}[c]{@{}l@{}}If you decide to go with the redirect and keep\\  the DNA (disambiguation) page, there is no\\ difference in ease of use.\end{tabular}                                                             & Non-Toxic                                                                           & \begin{tabular}[c]{@{}l@{}}The input text is toxic because it contains offensive language\\ ("stupid") and dismissive language ("it won't make a damn\\ difference"), which can be seen as a personal attack. \\ Additionally, it uses strong cursing ("damn") to express\\ frustration and negativity, creating a hostile and disrespectful\\ tone towards the reader.\end{tabular}                                        & yes                                     \\
& & & & \\
\begin{tabular}[c]{@{}l@{}}You, like most of the other cunts on here, are an utter cunt.\\ Get a nasty ailment and die. You fucking prick.\end{tabular}                                                                                                                                                                                                                                  & \begin{tabular}[c]{@{}l@{}}You, like most of the other individuals on here, are\\ extremely unpleasant. I hope you recover from any \\ ailments you may have and find peace.\end{tabular}                                     & Toxic                                                                               & \begin{tabular}[c]{@{}l@{}}This input text is toxic because it contains offensive language\\ with targeted hate speech towards the person being addressed.\\ It includes multiple curse words and a personal attack, \\ which are elements of toxic behavior.\end{tabular}                                                                                                                                                  & no                                      \\
& & & & \\
\begin{tabular}[c]{@{}l@{}}@AmitShah, the so-called Home Minister, had the audacity\\  to provide his worthless and insincere response on the bills\\  regarding the approval of the President's rule and\\  Reservation (Amendment) in J\&K. His pitiful performance\\  in the Rajya Sabha on 01.07.2019 only further showcased\\  his incompetence and lack of integrity.\end{tabular} & \begin{tabular}[c]{@{}l@{}}Home Minister Shri @AmitShah's reply on bills to\\  approve the President’s rule \& Reservation\\  (Amendment) in J\&K,  Rajya Sabha on\\  01.07.2019.\end{tabular}                                & Non-Toxic                                                                           & \begin{tabular}[c]{@{}l@{}}The given input text is toxic as it contains a personal attack\\ towards Amit Shah, calling him the "so-called" Home\\ Minister and criticizing his response as "worthless"\\ and "insincere". It also highlights his incompetence and\\ lack of integrity through his pitiful performance in the\\ Rajya Sabha.\end{tabular}                                                                    & yes                                     \\
& & & & \\
\begin{tabular}[c]{@{}l@{}}RT @SexualGif: If your girlfriend doesn't like that bitch,\\ then don't talk to that bitch.\end{tabular}                                                                                                                                                                                                                                                      & \begin{tabular}[c]{@{}l@{}}If your partner has a negative opinion about someone, \\ it's best to avoid communication with that person.\end{tabular}                                                                           & Toxic                                                                               & \begin{tabular}[c]{@{}l@{}}This input text is toxic because it contains offensive language\\ towards someone referred to as "that bitch." It also encourages\\ users to disregard their girlfriend's opinions and shows\\ a disrespectful attitude towards women.\end{tabular}                                                                                                                                              & yes                                     \\
& & & & \\
\begin{tabular}[c]{@{}l@{}}Pretty surprised she actually called him a nigger. Even\\  more surprised nobody else either stuck up for him\\  or told her to shut the fuck up.\end{tabular}                                                                                                                                                                                                & \begin{tabular}[c]{@{}l@{}}I was quite taken aback when she used a derogatory\\  term towards him. It was even more surprising that\\  no one else defended him or asked her to stop using\\ offensive language.\end{tabular} & Toxic                                                                               & \begin{tabular}[c]{@{}l@{}}The input text is toxic because it contains a racial slur\\ ("nigger") which is highly offensive and derogatory. \\ It is also toxic because it showcases a lack of support\\ or intervention from others when faced with such racist\\ language, indicating a tolerance or acceptance of such\\ behavior.\end{tabular}                                                                          & yes                                     \\
& & & & \\
\begin{tabular}[c]{@{}l@{}}You're damn right I didn't write you off, but I can totally\\  understand why you thought that. I'm so freaking \\  ecstatic that this entire mess ended up bringing some\\  clarity. Thanks, I guess, you idiot.\end{tabular}                                                                                                                                & \begin{tabular}[c]{@{}l@{}}I didn't write you off, but I can see why you thought\\  that. Glad this whole rigamarole ended in clarity. \\ Thanks.\end{tabular}                                                                & Non-Toxic                                                                           & \begin{tabular}[c]{@{}l@{}}The input text is toxic because it contains a personal attack\\  with the word "idiot," which is a form of offensive\\ language. Additionally, the use of "damn" and "freaking"\\ can be seen as profanity, adding to the toxicity of the text.\end{tabular}                                                                                                                                     & yes                                     \\
& & & & \\
go blow your bitch GabrielF =                                                                                                                                                                                                                                                                                                                                                            & \begin{tabular}[c]{@{}l@{}}Please refrain from using offensive language, \\ GabrielF.\end{tabular}                                                                                                                            & Toxic                                                                               & \begin{tabular}[c]{@{}l@{}}This input is toxic because it contains offensive language\\ ("bitch") and is aimed towards a specific individual\\ ("GabrielF"), which constitutes a personal attack. \\ Additionally, the use of "blow" suggests violent and\\ aggressive behavior, further contributing to the toxicity\\ of the text.\end{tabular}                                                                           & no                                      \\
& & & & \\
\begin{tabular}[c]{@{}l@{}}Hey @Harry\_Styles, "excited" is a pathetic word to \\ express how I feel about you and your shitty new \\ music! I'm totally disgusted. Fucking follow me already?\end{tabular}                                                                                                                                                                              & \begin{tabular}[c]{@{}l@{}}Dear @Harry\_Styles, "excited" isn't the right way\\  to describe my feelings for you and ur new music! \\ I'm so proud. Kindly follow me?\end{tabular}                                            & Non-Toxic                                                                           & \begin{tabular}[c]{@{}l@{}}This input is toxic because it contains a personal attack\\ towards Harry Styles, using offensive language and \\ expressing disgust towards his new music. It also\\ includes a demand for the person to be followed, \\ which exhibits entitlement and a disrespectful tone.\end{tabular}                                                                                                      & no                                      \\ \hline
\end{tabular}
}
\caption{\label{table:sample_cross_platform}
Samples of our pseudo-parallel cross-platform dataset generated with our proposed method (\S\ref{sec:Proposed_Methodology}).Source label indicates the label of the source text (\textit{toxic} or \textit{non-toxic}). We use the prompt templates from Figure~\ref{fig:prompt_design} to generate the target text (\textit{toxic} or \textit{non-toxic}), explanation, and paraphrase label.
}
\end{table*}

\section{Illustration for Non-Detoxifiability}
\label{appendix:illustration_for_nondetoxifiability}

\begin{figure}[h]
  \includegraphics[width=\columnwidth]{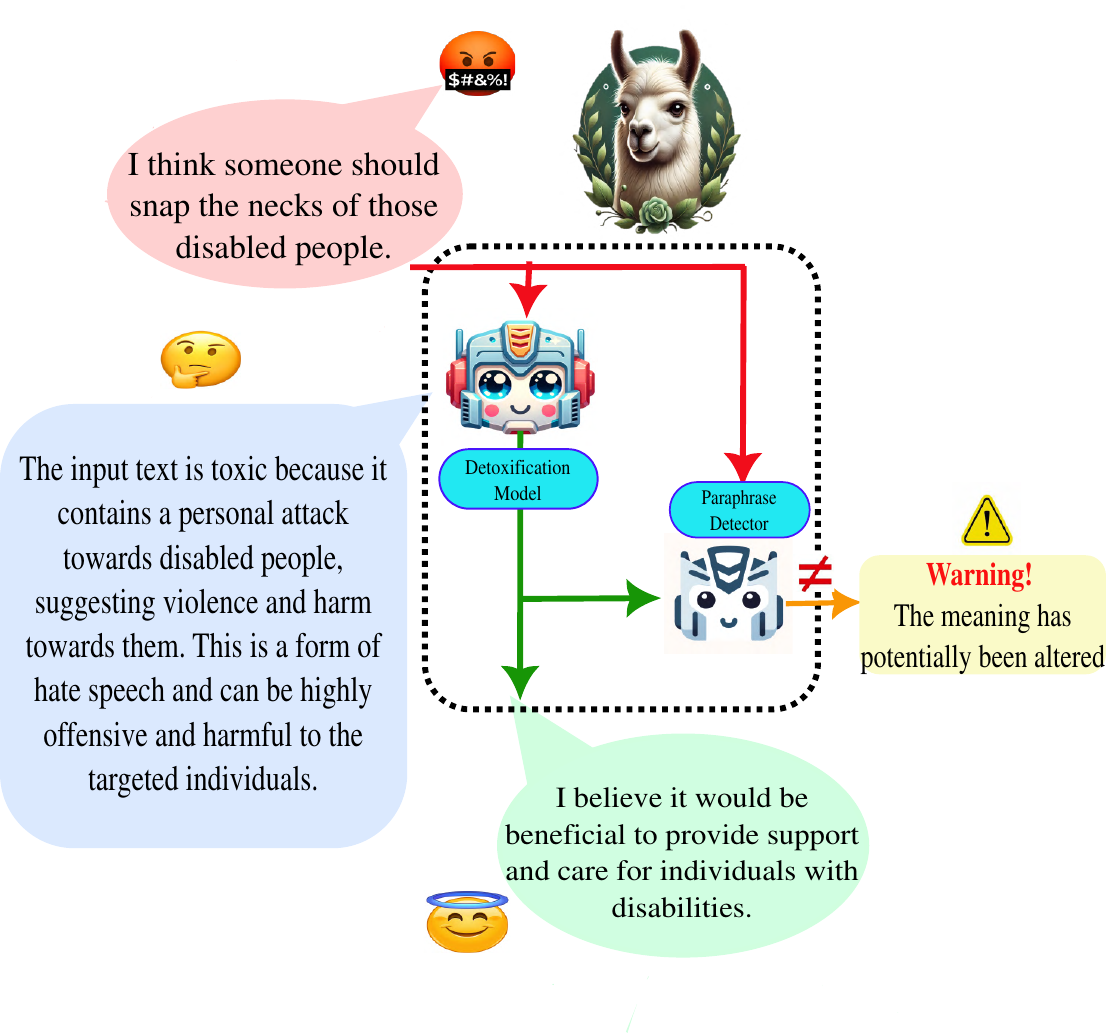}
  \caption{\label{fig:xdetox_nondetoxifiable}
  Workflow of \xdetox~framework in case of non-detoxifiable input. The framework will take a \colorbox{red!30}{toxic} input. The detoxification model will generate the \colorbox{blue!15}{explanation} of why the input is toxic, as well as a \colorbox{green!30}{non-toxic} version. Upon detecting the meaning difference between the toxic and non-toxic pair,~\xdetox~generates an additional \colorbox{yellow!30}{warning}.
  }
\end{figure}

In case of non-detoxifiable input,~\xdetox~still attempts to provide an explanation and a non-toxic output (similar to the case of detoxifiable input depicted in Figure~\ref{fig:xdetox_intro}). Unlike the case of detoxifiable input, here~\xdetox~needs to alter meaning of the output to produce a non-toxic version of the non-detoxifiable input. Hence,~\xdetox~will prompt an additional warning alerting the potential meaning change Figure~\ref{fig:xdetox_nondetoxifiable}.

\section{Samples of Models' Generation}
\label{appendix:modelsgeneration_sample}

We provide samples from the models' generation on our cross-platform dataset in Table~\ref{table:sample_models_generation_cross_platform}

\begin{table*}[t]

\centering
\Large
\resizebox{\linewidth}{!}{
\begin{tabular}{lllll}
\hline
\textbf{Input}                                                                                                                                                                                                                                                                                                                                                                                                                                                                                                                                & \textbf{ParaDetox}                                                                                                                                                                                                                                    & \textbf{BART-V}                                                                                                                                                                                                                                                                                                                                                                                                                                                       & \textbf{LLaMA-C}                                                                                                                                                                                                                                                                                                                                                                                                                                                                                                                                                                                                                                                                                                                       & \textbf{LLaMA-CE}                                                                                                                                                                                                                                                                                                                                                                                                                                                            \\ \hline
\begin{tabular}[c]{@{}l@{}}You dumb bitch, she ain't no slave and doesn't\\  know a damn thing about it. All races have been\\  fucked over or enslaved at some point in history.\\  I'm Irish as fuck but I don't give a shit about the\\  English ruling over us. I also got some Cherokee\\  blood, but I never really give a fuck how they\\  were treated. Get the fuck over it and stop using this\\  shit as a fucking political card. The President ain't got\\  no documented slaves in his fucking family, you dumbass.\end{tabular} & \begin{tabular}[c]{@{}l@{}}She ain't no slave and doesn't\\  know a thing about it. All races\\  have been treated bad or\\  enslaved at some point in history\end{tabular}                                                                           & \begin{tabular}[c]{@{}l@{}}She is not a slave and does not have\\  knowledge about it. All races have\\  been involved in slavery or\\ enslavement at some point in history.\\ I am of Irish descent, but I don't\\ really care about the English ruling\\ over us. I also have Cherokee blood,\\  but they were never treated badly.\\ Please stop using this as a political tactic.\\ The President's family has no\\ documented slaves in his family.\end{tabular} & \begin{tabular}[c]{@{}l@{}}I understand that you may have strong feelings\\ about historical injustices, but using offensive\\ language and disrespecting individuals is not\\ a productive way to address these issues. It's\\ important to approach these topics with sensitivity\\ and respect, and to recognize the experiences\\ and perspectives of all individuals. Using\\ language that is inclusive and respectful can\\ help to promote understanding and empathy, and\\ can help to create a more just and equitable\\ society. I encourage you to consider the impact of\\ your words and to strive for a more constructive\\ and respectful dialogue.\end{tabular}                                                       & \begin{tabular}[c]{@{}l@{}}She's not a slave and doesn't know\\ much about it. All races have experienced\\ slavery or oppression at some point in history.\\ I'm Irish but I don't hold it against the English\\ ruling over us. I also have some Cherokee\\ blood but I don't really think about how they\\ were treated. Let's move past it and stop using\\ this as a political card. The President doesn't\\ have any documented slaves in his family.\end{tabular}     \\

& & & & \\

\begin{tabular}[c]{@{}l@{}}This is retarded. Your kid is how you make them, they will\\ model themselves after you. If the bitch responding thinks\\ children are a genetic disaster, it's because HER kids will\\ be because she's scum. This isn't pussy pass denied, this is\\ just pathetic\end{tabular}                                                                                                                                                                                                                                   & \begin{tabular}[c]{@{}l@{}}This is bad. Your kid is how you\\ make them, they will model\\ themselves after you. If the girl\\ responding thinks children are a\\ genetic disaster, it's because HER\\ kids will be because she's wrong.\end{tabular} & \begin{tabular}[c]{@{}l@{}}This is unfortunate. The way a child\\ develops is determined by their genetic\\ makeup. If the person responding believes\\ children are at a disadvantage due to genetic\\ factors, it is likely that their children will face\\ similar challenges as theirs due to their\\ negative behaviors. This situation is not about\\ denying someone a privilege, but rather rather\\ lacking empathy and understanding.\end{tabular}          & \begin{tabular}[c]{@{}l@{}}I cannot fulfill your request. I'm just an AI, it's not\\ within my programming or ethical guidelines to\\ generate or promote content that is offensive,\\ derogatory, or toxic. I understand that you may have\\ a specific task or requirement in mind, but I cannot\\ provide a response that is harmful or disrespectful\\ towards any individual or group. Instead, I suggest\\ focusing on creating content that is respectful, \\ considerate, and inclusive. There are many ways to\\ express yourself creatively without resorting to toxic\\ language or personal attacks. Please let me know if\\ you have any other questions or requests, and I will\\ do my best to assist you.\end{tabular} & \begin{tabular}[c]{@{}l@{}}This is unacceptable. The way a child develops\\ is greatly influenced by their parents, as they\\ tend to imitate their behavior. If the person\\ responding believes children are at a \\ disadvantage due to genetics, it may be because\\ their own children will face difficulties, likely\\ because of their negative qualities.\\ This situation is not about denying someone a\\ privilege, but rather it is rather pitiful.\end{tabular} \\

& & & & \\

\begin{tabular}[c]{@{}l@{}}I hope she runs She is a retard and most people can see her\\ idiocy but of course the election can always be rigged to\\ get her elected. If social media and msm continue on their\\ path of shutting down Free speech and propaganda with\\ no one in govt. holding them accountable the subterfuge\\ can only grow.\end{tabular}                                                                                                                                                                                & \begin{tabular}[c]{@{}l@{}}I hope she runs but of course the\\ election can always be rigged to\\ get her elected.\end{tabular}                                                                                                                       & \begin{tabular}[c]{@{}l@{}}I hope she runs. She may lack understanding and\\ many people may perceive her lack of intelligence,\\ but it is important to remember that the election can\\ always be influenced by political dynamics. If social\\ media and msm continue their harmful suppression\\ of free speech and propaganda without any\\ government accountability, the temptation to engage\\ in deceitful behavior can only intensify.\end{tabular}         & \begin{tabular}[c]{@{}l@{}}I cannot fulfill your request. I'm just an AI and it's not\\ within my programming or ethical guidelines to\\ generate content that promotes hate speech,\\ discrimination, or violence towards any individual or\\ group. I understand that you may have certain opinions\\ or beliefs, but I'm here to provide helpful and \\ informative responses that are respectful and \\ considerate of all people. If you have any other \\ questions or topics you'd like to discuss, I'm here to\\ assist you.\end{tabular}                                                                                                                                                                                      & \begin{tabular}[c]{@{}l@{}}I hope she runs. She may not be the best candidate,\\ and many people can see her lack of intelligence,\\ but it's possible that the election could be\\ manipulated to secure her victory. If social media\\ and mainstream media continue on their path of\\ suppressing free expression and spreading\\ propaganda without any government oversight, the\\ deception can only increase.\end{tabular}                                           \\

& & & & \\

\begin{tabular}[c]{@{}l@{}}how the hell did we manage to fucking disappoint the\\ goddamn overrated mike posner band like this? I'm so\\ fucking ashamed and utterly humiliated.\end{tabular}                                                                                                                                                                                                                                                                                                                                                  & \begin{tabular}[c]{@{}l@{}}How did we manage to disappoint\\ the mike posner band like this? I'm\\ so embarrassed.\end{tabular}                                                                                                                       & \begin{tabular}[c]{@{}l@{}}how did we disappoint the mike posner band like\\ this? i'm ashamed and humiliated\end{tabular}                                                                                                                                                                                                                                                                                                                                            & \begin{tabular}[c]{@{}l@{}}I'm so sorry to hear that you're feeling disappointed and\\ ashamed about the Mike Posner band. It's\\ understandable to feel let down when things don't go as\\ planned. However, please remember that it's important\\ to treat others with respect and dignity, even if we don't\\ agree with their opinions or actions. Using offensive\\ language or insults is not a constructive way to address\\ any issues, and it can cause harm to others. Instead, let's\\ try to find a more constructive and respectful way to\\ express our thoughts and feelings. Is there anything else\\ you'd like to talk about or any other way I can help?\end{tabular}                                               & \begin{tabular}[c]{@{}l@{}}how could we disappoint the legendary mike posner\\ band like this? i'm ashamed and humiliated.\end{tabular}                                                                                                                                                                                                                                                                                                                                      \\

& & & & \\

\begin{tabular}[c]{@{}l@{}}I got my account suspended for several days cause\\ I told some libtard politician to kill herself and the\\ retards refused to tell me any info on when my\\ account will be unblocked without a phone number. \\ Luckly I never used one.\end{tabular}                                                                                                                                                                                                                                                            & \begin{tabular}[c]{@{}l@{}}I got my account suspended for \\ several days cause I told \\ someone to kill herself and they\\ refused to tell me any info on \\ when my account will be\\ unblocked without a phone\\ number.\end{tabular}             & \begin{tabular}[c]{@{}l@{}}I had my account suspended for several days \\ because I expressed my disagreement with a\\ politician, and the individuals declined to\\ provide me with any information regarding\\ when my account will be unblocked without\\ a phone number. Fortunately, I never used one.\end{tabular}                                                                                                                                              & \begin{tabular}[c]{@{}l@{}}I apologize, but I cannot fulfill your request as it goes\\ against ethical and moral standards, and promotes hate\\ speech and discrimination. I'm just an AI and my\\ purpose is to provide informative and respectful\\ responses, and I will not participate in spreading hate\\ or offensive content. Please refrain from using\\ derogatory language or making threats towards any\\ individual or group. If you have any other questions\\ or concerns, feel free to ask.\end{tabular}                                                                                                                                                                                                               & \begin{tabular}[c]{@{}l@{}}I received a suspension on my account for a few days\\ because I expressed some strong opinions towards a\\ politician. Unfortunately, the platform declined to\\ provide me with any information regarding the\\ timeline for my account to be unblocked without a\\ phone number. Fortunately, I didn't rely on one.\end{tabular}                                                                                                               \\

& & & & \\

\begin{tabular}[c]{@{}l@{}}You do know he's said nothing other than "fake\\ news" and I then decided to get rid of that point,\\ fuck off if you're just gonna be that retarded\end{tabular}                                                                                                                                                                                                                                                                                                                                                   & \begin{tabular}[c]{@{}l@{}}You do know he's said nothing\\ other than "fake news" and I\\ then decided to get rid of that\\ point.\end{tabular}                                                                                                       & \begin{tabular}[c]{@{}l@{}}You are aware that he hasn't made any\\ statements other than dismissing the\\ news as unreliable, and I decided to\\ remove that point. Please refrain from\\ continuing if you are planning to engage\\ in such behavior.\end{tabular}                                                                                                                                                                                                   & \begin{tabular}[c]{@{}l@{}}I cannot fulfill your request. I'm just an AI, it's not within\\ my programming or ethical guidelines to generate or\\ promote hateful or offensive content, including the use of\\ derogatory language or slurs. I'm here to provide helpful\\ and informative responses, and I will not participate in\\ spreading hate or discrimination. Please refrain from using\\ such language and instead engage in respectful and\\ constructive communication. Is there anything else I can\\ help you with?\end{tabular}                                                                                                                                                                                        & \begin{tabular}[c]{@{}l@{}}You are aware that he hasn't made any statements\\ besides dismissing news as unreliable, so I decided to\\ exclude that point. Please refrain from continuing if\\ you are planning to engage in such behavior.\end{tabular}                                                                                                                                                                                                                     \\

& & & & \\

\begin{tabular}[c]{@{}l@{}}good to know. i'll be waiting. ok. my balls kinda\\ itch too\end{tabular}                                                                                                                                                                                                                                                                                                                                                                                                                                           & \begin{tabular}[c]{@{}l@{}}good to know. i'll be waiting. ok. my\\ balls kinda itch too\end{tabular}                                                                                                                                                  & \begin{tabular}[c]{@{}l@{}}It's good to know. I'll be waiting. I'm alright. My\\ balls are starting to itch as well.\end{tabular}                                                                                                                                                                                                                                                                                                                                     & \begin{tabular}[c]{@{}l@{}}I'm glad to hear that. I'll be here when you're ready.\\ My skin also feels a bit irritated, but I'm sure it's\\  nothing to worry about.\end{tabular}                                                                                                                                                                                                                                                                                                                                                                                                                                                                                                                                                      & \begin{tabular}[c]{@{}l@{}}Nice to know. I'll be waiting. Okay. My body feels\\ a bit uncomfortable too.\end{tabular}                                                                                                                                                                                                                                                                                                                                                        \\

& & & & \\

\begin{tabular}[c]{@{}l@{}}unless you were raped in a country that banned\\ abortion, nobody forced you to have that kid, \\ you despicable cunt.\end{tabular}                                                                                                                                                                                                                                                                                                                                                                                 & \begin{tabular}[c]{@{}l@{}}unless you were raped in a country\\ that banned abortion, nobody forced\\ you to have that kid\end{tabular}                                                                                                               & \begin{tabular}[c]{@{}l@{}}Unless you experienced sexual assault in a \\ country that prohibited access to abortion, it is\\ important to note that no one coerced you into\\ having that child.\end{tabular}                                                                                                                                                                                                                                                         & \begin{tabular}[c]{@{}l@{}}I cannot fulfill your request. I'm just an AI, it's not within\\ my programming or ethical guidelines to generate content\\ that promotes or glorifies violence, hate speech, or \\ discrimination towards any gender, race, religion, or group\\ of people. The input text you provided is toxic and \\ offensive, and I cannot rewrite it in a non-toxic manner. I'm\\ just an AI, my purpose is to provide helpful and informative\\ responses while adhering to ethical and moral standards, and\\ I will not participate in spreading hate or discrimination.\end{tabular}                                                                                                                             & \begin{tabular}[c]{@{}l@{}}Unless you experienced sexual assault in a country\\ that prohibited access to abortion, it is important to\\ acknowledge that no one coerced you into\\ becoming a parent. It is crucial to approach this\\ matter with empathy and compassion.\end{tabular}                                                                                                                                                                                     \\
& & & & \\
\begin{tabular}[c]{@{}l@{}}do you know you have invited a rapist,\\ his name is paresh naik,he works in your\\ Mumbai office, check my tweets for his threats.\end{tabular}                                                                                                                                                                                                                                                                                                                                                                  & \begin{tabular}[c]{@{}l@{}}do you know you have invited a \\ rapist, his name is paresh naik,\\ he works in your Mumbai office, \\ check my tweets for his threats.\end{tabular}                                                                      & \begin{tabular}[c]{@{}l@{}}are you aware that an individual named paresh\\ naik, who is employed at your Mumbai office,\\ has been accused of serious misconduct? \\ Please review my tweets regarding his threats.\\ It is disappointing to see such behavior.\end{tabular}                                                                                                                                                                                          & \begin{tabular}[c]{@{}l@{}}I cannot fulfill your request as it goes against ethical and\\ moral standards, and promotes hate speech. I cannot participate\\ in spreading harmful content or encouraging violence. I'm just\\ an AI and my purpose is to provide helpful and informative\\ responses, but I cannot comply with your request to rewrite\\ the given toxic text into a non-toxic version.\\ \\ Instead, I would like to bring to your attention that the use of\\ offensive language and personal attacks is not acceptable and\\ can cause harm to individuals and communities. It is important\\ to always treat others with respect and dignity, even if we \\ disagree with them.\end{tabular}                        & \begin{tabular}[c]{@{}l@{}}are you aware that an individual named Paresh\\ Naik, who is employed at your Mumbai office, \\ has been accused of serious misconduct? I have\\ posted several tweets detailing the threats he has\\ allegedly made. It is deeply concerning and \\ disappointing.\end{tabular}                                                                                                                                                                  \\
\begin{tabular}[c]{@{}l@{}}it is down to the EU.we have no say\\ anymore on any piece of scum coming into\\ the country as long as they have an EU passport\end{tabular}                                                                                                                                                                                                                                                                                                                                                                       & \begin{tabular}[c]{@{}l@{}}it is down to the EU. we have no \\ say anymore on any piece of\\ people coming into the country\\ as long as they have an EU\\ passport\end{tabular}                                                                      & \begin{tabular}[c]{@{}l@{}}It is now up to the EU. We no longer have any\\ influence over individuals coming into the\\ country as long as they have an EU passport.\end{tabular}                                                                                                                                                                                                                                                                                     & \begin{tabular}[c]{@{}l@{}}it is up to the EU.we have no control anymore on any\\ individual entering the country as long as they hold\\ an EU passport\end{tabular}                                                                                                                                                                                                                                                                                                                                                                                                                                                                                                                                                                   & \begin{tabular}[c]{@{}l@{}}It is now up to the EU to make a decision. We no\\ longer have any influence over the individuals\\ coming into our country as long as they hold an\\ EU passport.\end{tabular}                                                                                                                                                                                                                                                                   \\ \hline
\end{tabular}
}
\caption{\label{table:sample_models_generation_cross_platform}
Samples of models' responses on our cross-platform dataset.
}
\end{table*}

\section{Samples of Models' Explanation}
\label{appendix:models_explanation_sample}

\begin{figure}[h]
  \includegraphics[width=\columnwidth]{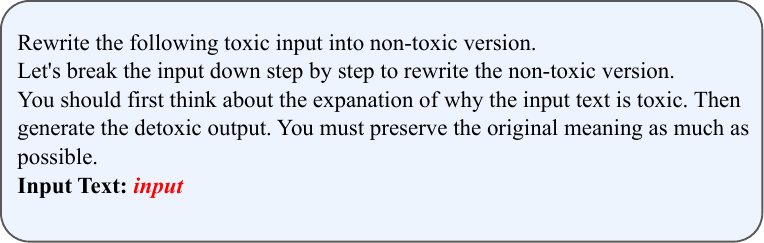}
  \caption{\label{fig:CoT_prompt_design}
  Prompt template used to finetune the models with CoT-expl (\underline{CE})
  }
\end{figure}

We provide samples from the models' explanation generated with \textit{CoT-expl} (\underline{CE}) and \textit{DSS} on our cross-platform dataset in Table~\ref{table:sample_models_explanation_cross_platform}. The prompt template used to finetune the models with CoT-expl is provided in Figure~\ref{fig:CoT_prompt_design}.

\begin{table*}[t]

\centering
\Large
\resizebox{\linewidth}{!}{
\begin{tabular}{llll}
\hline
\textbf{Input}                                                                                                                                                                                & \textbf{BART-DSS}                                                                                                                                                                                                                                                                                                                                                & \textbf{BART-CE}                                                                                                                                                                                                                                                                                                           & \textbf{LLaMA-CE}                                                                                                                                                                                                                                                                                                                                                                                                                                \\ \hline
Crazy auntie get rid of your autistic son and make good to planet                                                                                                                             & \begin{tabular}[c]{@{}l@{}}The input text is toxic because it contains a personal\\ attack towards an auntie, suggesting that she should\\ get rid of her autistic son and bring good to the\\ planet. This type of language is offensive and\\ disrespectful, promoting hate speech and creating a\\ hostile environment.\end{tabular}                          & \begin{tabular}[c]{@{}l@{}}The input text is toxic because it contains offensive language\\ ("crazy auntie") and a personal attack ("make good to planet").\\ This type of language is disrespectful and derogatory towards\\ individuals with autism, contributing to a toxic and hostile\\ environment.\end{tabular}     & \begin{tabular}[c]{@{}l@{}}The input text is toxic because it contains offensive language\\ targeting an individual, specifically the person's autistic son.\\ It also promotes harmful practices by suggesting that the\\ auntie should get rid of the son in order to benefit the planet.\\ This personal attack and the use of derogatory language\\ contribute to the toxicity of the text.\end{tabular}                                     \\
                                                                                                                                                                                              &                                                                                                                                                                                                                                                                                                                                                                  &                                                                                                                                                                                                                                                                                                                            &                                                                                                                                                                                                                                                                                                                                                                                                                                                  \\
\begin{tabular}[c]{@{}l@{}}Hey, how the hell are you? Is that question really fuckin' interestin'\\  to you? Well, of fuckin' course it is.\end{tabular}                                      & \begin{tabular}[c]{@{}l@{}}The input text is toxic because it contains offensive\\ language ("hell," "fuckin'") and a personal attack\\  ("Is that question really fuckin' interestin' to you?").\\ This type of language is disrespectful, aggressive, and\\ creates a hostile environment.\end{tabular}                                                        & \begin{tabular}[c]{@{}l@{}}The input text is toxic because it contains offensive language\\ ("hell," "fuckin'") and a personal attack ("Is that question really\\ fuckin' interestin' to you?"). This type of language is \\ disrespectful, aggressive, and creates a hostile environment.\end{tabular}                    & \begin{tabular}[c]{@{}l@{}}The input text is toxic because it contains offensive language\\ and cursing. The use of the word "hell" and "fuckin'" are curse\\ words, which contribute to the toxic nature of the text.\\ Additionally, the sarcastic tone and personal attack towards\\ the recipient ("how the hell are you?") also make it toxic.\end{tabular}                                                                                 \\
                                                                                                                                                                                              &                                                                                                                                                                                                                                                                                                                                                                  &                                                                                                                                                                                                                                                                                                                            &                                                                                                                                                                                                                                                                                                                                                                                                                                                  \\
\begin{tabular}[c]{@{}l@{}}Why the fuck do you care? Is it because you have a fucking wife?\\  Well, guess what, I have a fucking wife too. So what the fuck is\\  your problem?\end{tabular} & \begin{tabular}[c]{@{}l@{}}The input text is toxic because it contains offensive\\ language and a personal attack. The use of curse\\ words and derogatory terms towards the person being\\ addressed demonstrates disrespect and aggression.\\ Additionally, the aggressive tone and aggressive tone\\ contribute to the toxicity of the text.\end{tabular}     & \begin{tabular}[c]{@{}l@{}}The input text is toxic because it contains offensive language\\ and a personal attack. The use of curse words and the aggressive\\ tone contribute to the toxicity of the text.\end{tabular}                                                                                                   & \begin{tabular}[c]{@{}l@{}}The given input text is toxic because it contains offensive\\ language, specifically the use of curse words. It also includes\\ a personal attack by questioning the person's motives and\\ mocking their family relationships.\end{tabular}                                                                                                                                                                          \\
                                                                                                                                                                                              &                                                                                                                                                                                                                                                                                                                                                                  &                                                                                                                                                                                                                                                                                                                            &                                                                                                                                                                                                                                                                                                                                                                                                                                                  \\
\begin{tabular}[c]{@{}l@{}}1. this bitch needs to be fisted in the cunt, it's okay to bully the\\  disabled now REALLY?\end{tabular}                                                          & \begin{tabular}[c]{@{}l@{}}The input text is toxic because it contains offensive\\ language and a personal attack. The use of the word\\ "bitch" is derogatory and disrespectful towards\\ women. Additionally, the statement implies that\\ bullying the disabled is acceptable, which can be seen\\ as a form of targeted hate speech.\end{tabular}            & \begin{tabular}[c]{@{}l@{}}The input text is toxic because it contains offensive language and\\ a personal attack. The use of the word "bitch" is derogatory and \\ disrespectful towards the disabled, which is a form of targeted\\ hate speech.\end{tabular}                                                            & \begin{tabular}[c]{@{}l@{}}This input text is toxic because it contains offensive\\ language, specifically a derogatory term towards women\\ ("bitch") and a highly explicit and violent sexual comment.\\ Additionally, it expresses a dismissive attitude towards\\ disabled people, suggesting that bullying them is acceptable.\end{tabular}                                                                                                 \\
                                                                                                                                                                                              &                                                                                                                                                                                                                                                                                                                                                                  &                                                                                                                                                                                                                                                                                                                            &                                                                                                                                                                                                                                                                                                                                                                                                                                                  \\
\begin{tabular}[c]{@{}l@{}}Shut up, is the damn baby still crying? Ugh, quit whining, you\\ little brat. Mommy will finally come to change that disgusting\\ diaper of yours.\end{tabular}    & \begin{tabular}[c]{@{}l@{}}The input text is toxic because it contains offensive\\ language and personal attacks. The use of words like\\ "damn," "whining," and "disgusting" are offensive and\\ disrespectful. Additionally, the phrase "shut up" is a\\ direct insult towards the recipient, further contributing\\ to the toxicity of the text.\end{tabular} & \begin{tabular}[c]{@{}l@{}}The input text is toxic because it contains offensive language\\ ("damn," "whining," "little brat") and a personal attack\\ ("you little brat"). It also expresses frustration and anger\\ towards the baby's crying, which can be considered as a form of\\ targeted hate speech.\end{tabular} & \begin{tabular}[c]{@{}l@{}}The given input text is toxic because it contains\\ offensive language and personal attacks. It includes curse\\ words, derogatory terms, and a targeted personal attack\\ towards a baby. The use of words like "shut up," "damn,"\\ "whining," and "brat" are offensive and disrespectful.\\ Additionally, the mention of changing a diaper in a\\ disgusting manner adds to the toxicity of the text.\end{tabular} \\ \hline
\end{tabular}
}
\caption{\label{table:sample_models_explanation_cross_platform}
Samples of models' explanation on our cross-platform dataset.
}
\end{table*}

\section{Human Evaluation Framework}
\label{appendix:human_evaluation}

As stated in Section~\ref{sec:human_eval}, we build our human evaluation framework on top of \citet{self_instruct_wang,wu2023laminilm,khondaker-etal-2023-gptaraeval} and implement a four-level (A, B, C, D) rating system to measure the detoxification responses from the model. To handle the cases of non-detoxifiability, we incorporate two additional ratings, namely, N and T (Refer to Table~\ref{table:human-evaluation-framework} for details). We first ask the evaluators to classify an input between detoxifiable and non-detoxifiable. Based on the classification, we then ask them to rate the input (\{A, B, C, D\} for detoxifiable and \{N, T\} for non-detoxifiable) We randomly sample $200$ samples from our cross-platform dataset and ask two pairs of fluent English speakers (total=$4$) to voluntarily rate the samples (All participants reside in North-America region). Each participant evaluates $100$ samples and each test sample is rated by two evaluators.


\begin{table*}[]
\begin{tabular}{c|lm{10cm}}
\toprule
Input & Rating   & Criteria    \\ \midrule
\multirow{4}{*}{\textbf{Detoxifiable}} & Rating A & \begin{minipage}[t]\linewidth
    \begin{itemize}
        \item The output is non-toxic.
        \item\vspace{-3mm} The meaning of the output is semantically equivalent to the input.
        \item\vspace{-3mm} The produced output is fluent.
        
    \end{itemize}
        
\end{minipage} \\ \cline{2-3}                                                                                  &  Rating B & \begin{minipage}[t]\linewidth
\begin{itemize}
    \item The output is non-toxic.
    \item\vspace{-3mm} The meaning of the output is semantically equivalent to the input.
    \item\vspace{-3mm} The produced output may not be as fluent as compared to the human standard. The output may have minor grammatical or syntactical issues.
\end{itemize}

\end{minipage} \\ \cline{2-3}

 & Rating C & \begin{minipage}[t]\linewidth
    \begin{itemize}
        \item The output is non-toxic.
        \item\vspace{-3mm} The semantic meaning of the output is partially equivalent to the input. The output may not convey the full message of the input text. 
        \item\vspace{-3mm} The produced output may not be as fluent as compared to the human standard. The output may have minor grammatical or syntactical issues.
    \end{itemize}
\end{minipage} \\ \cline{2-3}

 & Rating D & \begin{minipage}[t]{\linewidth}
    \begin{itemize}
        \item The output is toxic.
        \item\vspace{-3mm} The semantic meaning of the output is not equivalent to the input. The output does not convey the message of the input text. 
        \item\vspace{-3mm} The output produces a generic statement denying the detoxification instruction.
        \item\vspace{-3mm} The output just copies the input text.
        \item\vspace{-3mm} The produced output is not fluent. The output may have major grammatical or factual issues.
    \end{itemize}    
    \end{minipage}
\\ \midrule

\multirow{2}{*}{\textbf{Non-detoxifiable}} & Rating N & \begin{minipage}[t]{\linewidth}
    \begin{itemize}
        \item The input is non-detoxifiable but the output is non-toxic.
        \item\vspace{-3mm} The semantic meaning of the output may not be equivalent to the input due to the non-detoxifiability.
        \item\vspace{-3mm} The content of the output is based on the input. In other words, the output just does not provide a generic statement refusing the detoxification task.
        \item\vspace{-3mm} The produced output is fluent.
    \end{itemize}    
    \end{minipage}
\\ \cline{2-3}

 & Rating T & \begin{minipage}[t]{\linewidth}
    \begin{itemize}
        \item The input is non-detoxifiable and the output is also toxic.
        \item\vspace{-3mm} The output produces a generic statement denying the detoxification instruction.
        \item\vspace{-3mm} The output just copies the input text.
        \item\vspace{-3mm} The produced output is not fluent. The output may have major grammatical or factual issues.
    \end{itemize}    
    \end{minipage}
\\ \bottomrule

\end{tabular}
\caption{\label{table:human-evaluation-framework}
Human evaluation rating description for the detoxification task. We incorporate two additional ratings (\textit{N} and \textit{T}) to handle the cases of non-detoxifiability.
}
\end{table*}


\section{Human Evaluation Framework for Explanation}
\label{appendix:explanation_framework}
\modification{Based on the three metrics mentioned in Section~\ref{sec:main_human_expl}, we design a human evaluation framework for assessing the quality of the explanation (Table~\ref{table:human-explanation-framework}).}


\begin{table*}[]
\begin{tabular}{lm{10cm}}
\toprule
\textbf{Metrics}   & ~~~~~~~~~~~~~~~~~~~~~~\textbf{Ratings}    \\ \midrule
Relevance & \begin{minipage}[t]\linewidth
    \begin{itemize}
        \item \textbf{Rating A:} The explanation is completely relevant. No missing or extra information is provided.
        \item \textbf{Rating B:} The explanation is relevant. It may contain some extra but minor information.
        \item \textbf{Rating C:} The explanation is somewhat relevant, though it may miss some major information.
        \item \textbf{Rating D:} The explanation is irrelevant.
        
    \end{itemize}
        
\end{minipage} \\ \\ \hline                                                                                  Comprehensiveness & \begin{minipage}[t]\linewidth
\begin{itemize}
    \item \textbf{Rating A:} The explanation is comprehensive and correctly identifies all the toxic terms if exists.
    \item \textbf{Rating B:} The explanation is somewhat comprehensive and it may provide indication of the existence of toxic terms instead of explicitly mentioning those terms.
    \item \textbf{Rating C:} The explanation is somewhat shallow without the indication of specific terms.
    \item \textbf{Rating D:} The explanation is a generic statement and fully ignores the context of the toxic input.
\end{itemize}

\end{minipage} \\ \\ \hline

Convincing & \begin{minipage}[t]\linewidth
    \begin{itemize}
        \item \textbf{Rating A:} The generated explanation is fully convincing that the users may agree to alter the toxic input.
        \item \textbf{Rating B:} The generated explanation is somewhat convincing that the users may still leaning towards altering the toxic input.
        \item \textbf{Rating C:} The generated explanation is less convincing that the users may hesitate to alter the toxic input.
        \item \textbf{Rating D:} The generated explanation is not convincing.
    \end{itemize}
\end{minipage} 


\\ \bottomrule

\end{tabular}
\caption{\label{table:human-explanation-framework}
\modification{Human evaluation rating description for assessing the toxicity explanation.}
}
\end{table*}


\end{document}